\documentclass{article} 
\usepackage{iclr2024_conference,times}

\usepackage{amsmath,amsfonts,bm}

\def\eqref#1{equation~\ref{#1}}

\def\1{\bm{1}}

\def\ervb{{\textnormal{b}}}
\def\ervc{{\textnormal{c}}}

\def\ervf{{\textnormal{f}}}

\def\ermS{{\textnormal{S}}}

\def\vh{{\bm{h}}}

\def\vx{{\bm{x}}}

\def\vzfd{{\bm{z}_\text{\rm FD}}}

\def\mA{{\bm{A}}}

\def\mL{{\bm{L}}}

\def\mQ{{\bm{Q}}}

\DeclareMathAlphabet{\mathsfit}{\encodingdefault}{\sfdefault}{m}{sl}
\SetMathAlphabet{\mathsfit}{bold}{\encodingdefault}{\sfdefault}{bx}{n}

\def\gG{{\mathcal{G}}}

\usepackage{hyperref}
\usepackage{url}

\title{Causal Inference with \\Conditional Front-Door Adjustment and\\Identifiable Variational Autoencoder}

\iclrfinalcopy

\author{Ziqi Xu$^1$, Debo Cheng$^1$, Jiuyong Li$^1$, Jixue Liu$^1$,  Lin Liu$^1$ \& Kui Yu$^2$  \\
$^1$University of South Australia\\
$^2$Hefei University of Technology\\
}

\usepackage{array}
\usepackage{amsfonts}       
\usepackage{amsthm}
\usepackage{amsmath}
\usepackage{theoremref}
\usepackage{xcolor}
\usepackage{titlesec}
\usepackage{multirow}
\usepackage{diagbox}
\usepackage{threeparttable}
\usepackage{verbatim}
\usepackage{booktabs}
\usepackage{subcaption}
\usepackage{setspace}
\usepackage{graphicx}     
\usepackage{nicefrac}       
\usepackage{microtype}            
\usepackage{wrapfig}
\usepackage{algorithm}
\usepackage{algorithmic}
\usepackage{amssymb}
\usepackage{enumitem}

\newtheorem{definition}{Definition}
\newtheorem{assumption}{Assumption}

\newtheorem*{problem}{Model Setting}

\newtheorem{theorem}{Theorem}

\makeatletter
\newcommand*{\indep}{%
	\mathbin{%
		\mathpalette{\@indep}{}%
	}%
}
\newcommand*{\nindep}{%
	\mathbin{
		\mathpalette{\@indep}{\not}
	}%
}
\newcommand*{\@indep}[2]{%
	\sbox0{$#1\perp\m@th$}
	\sbox2{$#1=$}
	\sbox4{$#1\vcenter{}$}
	\rlap{\copy0}
	\dimen@=\dimexpr\ht2-\ht4-.2pt\relax
	\kern\dimen@
	{#2}%
	\kern\dimen@
	\copy0 %
}
\newcommand\inv[1]{#1\raisebox{1.15ex}{$\scriptscriptstyle-\!1$}}

\begin{document}

\maketitle

\begin{abstract}
	An essential and challenging problem in causal inference is causal effect estimation from observational data. The problem becomes more difficult with the presence of unobserved confounding variables. The front-door adjustment is a practical approach for dealing with unobserved confounding variables. However, the restriction for the standard front-door adjustment is difficult to satisfy in practice. In this paper, we relax some of the restrictions by proposing the concept of conditional front-door (CFD) adjustment and develop the theorem that guarantees the causal effect identifiability of CFD adjustment. Furthermore, as it is often impossible for a CFD variable to be given in practice, it is desirable to learn it from data. By leveraging the ability of deep generative models, we propose CFDiVAE to learn the representation of the CFD adjustment variable directly from data with the identifiable Variational AutoEncoder and formally prove the model identifiability. Extensive experiments on synthetic datasets validate the effectiveness of CFDiVAE and its superiority over existing methods. The experiments also show that the performance of CFDiVAE is less sensitive to the causal strength of unobserved confounding variables. We further apply CFDiVAE to a real-world dataset to demonstrate its potential application.
\end{abstract}

\section{Introduction}
Estimating causal effects is a fundamental problem in many application areas. For example, policymakers need to know whether the implementation of a policy has a positive impact on the community~\citep{athey2017beyond,tran2022most}, and medical researchers study the effects of treatments on patients~\citep{petersen2014causal}. Randomised Controlled Trials (RCTs)~\citep{fisher1936design} are considered the golden standard for estimating causal effects. However, RCTs are difficult to implement in many real-world cases due to ethical issues or high costs~\citep{deaton2018understanding}. For example, it would be unethical to subject an individual to a condition (e.g., smoking) if the condition may have potentially negative consequences. Therefore, many methods have been developed to infer causal effects from observational data. Most of the methods assume no unobserved variables affecting both the treatment and outcome, i.e., the unconfoundedness assumption~\citep{imbens2015causal}, and follow the back-door criterion~\citep{pearl2009causality} to determine valid adjustment variable for unbiased estimation.

\begin{figure}
	\vspace{-0.5cm}
	\centering
	\begin{subfigure}[b]{0.20\textwidth}
		\centering
		\includegraphics[scale=0.3]{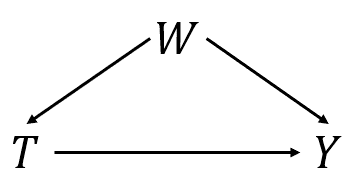}
		\caption{}
		\label{pic:intro1}
	\end{subfigure}
	\begin{subfigure}[b]{0.28\textwidth}
		\centering
		\includegraphics[scale=0.3]{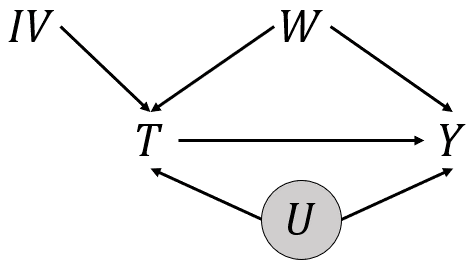}
		\caption{}
		\label{pic:intro2}
	\end{subfigure}
	\begin{subfigure}[b]{0.24\textwidth}
		\centering
		\includegraphics[scale=0.3]{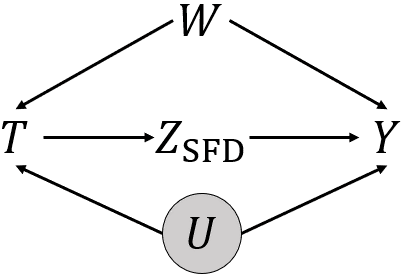}
		\caption{}
		\label{pic:intro3}
	\end{subfigure}
	\begin{subfigure}[b]{0.24\textwidth}
		\centering
		\includegraphics[scale=0.3]{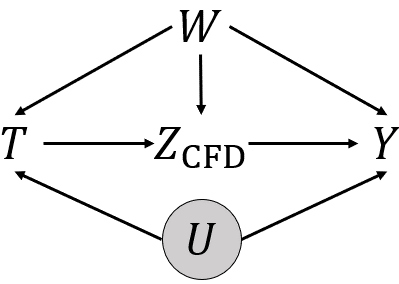}
		\caption{}
		\label{pic:intro4}
	\end{subfigure}
	\caption{Typical cases in causal effect estimation. $T$ is the treatment; $Y$ is the outcome; $W$ is the observed confounding variable; $U$ is the unobserved confounding variable; $IV$ is the instrumental variable; $Z_{\text{SFD}}$ is the standard front-door adjustment variable; and $Z_{\text{CFD}}$ is the conditional front-door adjustment variable}
	\label{pic:intro}
	\vspace{-0.2cm}
\end{figure}

A graphical view of the typical cases in causal effect estimation is shown in Fig.~\ref{pic:intro}. A simple case that satisfies the unconfoundedness assumption is illustrated in Fig.~\ref{pic:intro1}. In this case, the causal effect can be unbiasedly estimated by back-door adjustment~\citep{pearl2009causality}. Fig.~\ref{pic:intro2}, Fig.~\ref{pic:intro3} and Fig.~\ref{pic:intro4} show three cases where the unconfoundedness assumption is not satisfied. The IV (instrumental variable) approach has been extensively studied and commonly used to deal with the case shown in Fig.~\ref{pic:intro2}. However, in practice, IV is not always available. In this case, if there exists a standard front-door adjustment variable, e.g., $Z_{\text{SFD}}$ as indicated in Fig.~\ref{pic:intro3}, the standard front-door adjustment provides an effective approach to dealing with unobserved confounding variables. 

However, the requirement for a valid standard front door adjustment variable is too strict, which hinders their practical application. In this paper, we aim to relax the requirement by considering a more practical setting as shown in Fig.~\ref{pic:intro4}. Different from the standard front-door adjustment setting in Fig.~\ref{pic:intro3}, we allow the interaction between observed confounding variable ($W$) and the mediator ($Z_{\text{CFD}}$), and we call $Z_{\text{CFD}}$ a \emph{conditional} front-door (CFD) adjustment variable. This is a more practical setting. For instance, referring to Fig.~\ref{pic:intro4}, smoking ($T$) does not directly affect lung cancer development ($Y$) but mediated through tar in lungs ($Z_{\text{CFD}}$). For each patient, their other attributes such as age ($W$) can directly affect smoking, tar in lungs and lung cancer development. In this case, the standard front-door adjustment cannot be used since $Z_{\text{\rm CFD}}$ is no longer a standard front-door adjustment variable because it does not meet the standard front-door criterion (Definition~\ref{def:FD}), since, there is an unblocked back-door path from $T$ to $Z_{\text{\rm CFD}}$ ($T \gets W \to Z_{\text{\rm CFD}}$), and a back-door path from $Z_{\text{\rm CFD}}$ to $Y$ ($Z_{\text{\rm CFD}} \gets W \to Y$) which is not blocked by $T$.  

Additionally, it is unrealistic to assume that users always know a CFD adjustment variable in advance and thus it is desirable to find a CFD adjustment variable from observational data. In this paper, we propose a novel method, CFDiVAE, which is based on the identifiable VAE technique~\citep{KhemakhemKMH20} to learn the representation of a latent CFD variable from its proxy. We consider it is practical to assume the existence of proxies of a CFD adjustment variable. For instance, in the above example, the investigator may not observe tar in patients' lungs but they may observe the proxy variables, such as the results of patients' follow-up sputum tests and urine tests.

This paper advances the theory and practical use of causal inference in the presence of unobserved confounding variables through the following contributions:
\begin{itemize}[leftmargin=0.5cm]
	\item We identify and study a practical but challenging case of causal effect estimation when there exist unobserved confounding variables and the standard front-door adjustment is no longer applicable. We propose and formally define the concept of conditional front-door (CFD) adjustment and provide the theoretical guarantee of the causal effect identifiability of CFD adjustment. 
	\item We propose a novel model, CFDiVAE, to learn the representation of a CFD adjustment variable directly from observational data for unbiased average treatment effect estimation. We further provide the theoretical guarantee of the identifiability of the CFDiVAE model.
	\item We evaluate the effectiveness of CFDiVAE on both synthetic and real-world datasets. Experiments with synthetic datasets show that CFDiVAE outperforms existing methods. Furthermore, we apply CFDiVAE to a real-world dataset to show the application scenarios and potential of CFDiVAE.
\end{itemize}

\section{Preliminaries}
In this section, we present the necessary background of causal inference. We use a capital letter to represent a variable and a lowercase letter to represent its value. Boldfaced capital and lowercase letters are used to represent sets of variables and values, respectively. 

Let $\displaystyle \gG = (\mathbf{V}, \mathbf{E})$ be a directed acyclic graph (DAG), where $\mathbf{V}$ is the set of nodes and $\mathbf{E}$ is the set of edges between the nodes.
\begin{assumption} [Markov Condition~\citep{pearl2009causality}]
	\label{Markovcondition}
	Given a DAG $\mathcal{G}=(\mathbf{V}, \mathbf{E})$ and $P(\mathbf{V})$, the joint probability distribution of $\mathbf{V}$, $\mathcal{G}$ satisfies the Markov Condition if $\forall V_i \in \mathbf{V}$, $V_i$ is probabilistically independent of all of its non-descendants, given $Pa(V_i)$, the set of all parent nodes of $V_i$.
\end{assumption}
\begin{assumption}[Faithfulness~\citep{spirtes2000causation}]
	\label{Faithfulness}
	A DAG {$\mathcal{G}=(\mathbf{V}, \mathbf{E})$} is faithful to {$P(\mathbf{V})$} iff every conditional independence present in {$P(\mathbf{V})$} is entailed by {$\mathcal{G}$} and satisfies the Markov Condition. {$P(\mathbf{V})$} is faithful to {$\mathcal{G}$} iff there exists {$\mathcal{G}$} which is faithful to {$P(\mathbf{V})$}.
\end{assumption}

When the Markov condition and faithfulness assumption are satisfied, we can use $d$-separation to read the conditional independence between variables entailed in the DAG $\mathcal{G}$. Due to page limitation, we provide the definitions of causal path, non-causal path, $d$-separation and $d$-connect in Appx.~\ref{sec:Causality}.

This paper is focused on estimating the average treatment effect as defined below.
\begin{definition}[Average Treatment Effect (ATE)]
	\label{def:ATE}
	The average treatment effect of a treatment, denoted as $T$, on the outcome of interest, denoted as $Y$, is defined as $ATE=\mathbb{E}(Y \mid do(T=1))-\mathbb{E}(Y \mid do(T=0))$, where $do()$ is the $do$-operator and $do(T=t)$ represents the manipulation of the treatment by setting its value to $t$~\citep{pearl2009causality}.
\end{definition} 

When the context is clear, we abbreviate $do(T=t)$ as $do(t)$. In order to allow the above $do()$ expressions to be recovered from data, Pearl formally defined causal effect identifiability~\citep{pearl2009causality} (p.77) and proposed two well-known identification conditions, the back-door criterion and front-door criterion.
\begin{definition}[Back-Door Criterion~\citep{pearl2009causality}]
	\label{def:BD}
	A set of variables $Z_{\text{\rm BD}}$ satisfies the back-door criterion relative to an ordered pair of variables $(T,Y)$ in a DAG $\mathcal{G}$ if: (1) no node in $Z_{\text{\rm BD}}$ is a descendant of $T$; and (2) $Z_{\text{\rm BD}}$ blocks every path between $T$ and $Y$ that contains an arrow into $T$.
\end{definition}

A back-door path is a non-causal path from $T$ to $Y$. They have been recognised as ``back-door” paths because they flow backwards out of $T$, i.e., a back-door path points into $T$.

\begin{theorem}[Back-Door Adjustment~\citep{pearl2009causality}]
	If $Z_{\text{\rm BD}}$ satisfies the back-door criterion relative to $(T,Y)$, then the causal effect of $T$ on $Y$ is identifiable and is given by the following back-door adjustment formula~\citep{pearl2009causality}:
	\begin{equation}
		\begin{aligned}
			P(y|do(t)) = \sum_{z_{\text{\rm BD}}}^{}P(y\mid t,z_{\text{\rm BD}})P(z_{\text{\rm BD}}).
		\end{aligned}	
	\end{equation}
\end{theorem}
\begin{definition}[Front-Door Criterion~\citep{pearl2009causality}]
	\label{def:FD}
	A set of variables $Z_{\text{\rm SFD}}$ is said to satisfy the (standard) front-door criterion relative to an ordered pair of variables $(T,Y)$ in a DAG $\mathcal{G}$ if: (1) $Z_{\text{\rm SFD}}$ intercepts all directed paths from $T$ to $Y$; (2) there is no unblocked back-door path from $T$ to $Z_{\text{\rm SFD}}$; and (3) all back-door paths from $Z_{\text{\rm SFD}}$ to $Y$ are blocked by $T$.
\end{definition}

\begin{theorem}[Front-Door Adjustment~\citep{pearl2009causality}]
	If $Z_{\text{\rm SFD}}$ satisfies the (standard) front-door criterion relative to $(T,Y)$, then the causal effect of $T$ on $Y$ is identifiable and is given by the following standard front-door adjustment formula~\citep{pearl2009causality}:
	\begin{equation}
		\label{eqa:FD2009}
		\begin{aligned}
			P(y|do(t)) = \sum_{z_{\text{\rm SFD}},t'}^{}P(y\mid t',z_{\text{\rm SFD}})P(t')P(z_{\text{\rm SFD}}\mid t),
		\end{aligned}
	\end{equation}where $t'$ is a distinct realisation of treatment.
\end{theorem}

\section{Conditional Front-Door Adjustment}
\label{sec:Identifiability}
In this section, we present the definition of conditional front-door criterion and the theorem showing that the average causal effect of treatment $T$ on outcome $Y$ is identifiable via conditional front-door adjustment. The causal effect of $T$ on $Y$ is identifiable if the quantity $p(y\mid do(t))$ can be computed uniquely from any positive probability of the observed variables~\citep{pearl2009causality}. We formally define the conditional front-door criterion as follows:

\begin{definition}[Conditional Front-Door (CFD) Criterion]
	\label{def:FD1}
	A set of variables $Z_{\text{\rm CFD}}$ is said to satisfy the conditional front-door criterion relative to an ordered pair of variables $(T,Y)$ in a DAG $\mathcal{G}$ if: (1) $Z_{\text{\rm CFD}}$ intercepts all directed paths from $T$ to $Y$; (2) there exists a set of variables $W$, called the conditioning variables of $Z_{\text{\rm CFD}}$, such that all back-door paths from $T$ to $Z_{\text{\rm CFD}}$ are blocked by $W$; and (3) all back-door paths from $Z_{\text{\rm CFD}}$ to $Y$ are blocked by $\{T\} \cup W$.
\end{definition}
Fig.~\ref{pic:intro4} provides an illustration of CFD criterion, where $Z_{\text{\rm CFD}}$ satisfies the criterion, and $W$ is the conditioning variable of $Z_{\text{\rm CFD}}$. The following theorem provides the theoretical guarantee of the identifiability of the causal effect of $T$ on $Y$ via CFD adjustment and gives the adjustment formula. 

\begin{theorem}[Conditional Front-Door (CFD) Adjustment]
	\label{def:FD2}
	If $Z_{\text{\rm CFD}}$ satisfies the CFD criterion relative to $(T,Y)$, the causal effect of $T$ on $Y$ is identifiable and is given by the following CFD adjustment formula:
	\begin{equation}
		\label{eqa:FD1}
		\begin{aligned}
			P(y|do(t)) = \sum_{z_{\text{\rm CFD}},w,t'} P(y \mid t',z_{\text{\rm CFD}},w) P(t' \mid w) P(z_{\text{\rm CFD}} \mid t,w) P(w),
		\end{aligned}
	\end{equation}where $t'$ is a distinct realisation of treatment.
\end{theorem}
Proof of the above theorem is provided in Appx.~\ref{sec:proofFD2}.

\section{The Proposed CFDiVAE Model}
\subsection{Problem Setup}
\begin{wrapfigure}{r}{0.31\textwidth}
	\vspace{-1.8cm}
	\centering
	\includegraphics[scale=0.35]{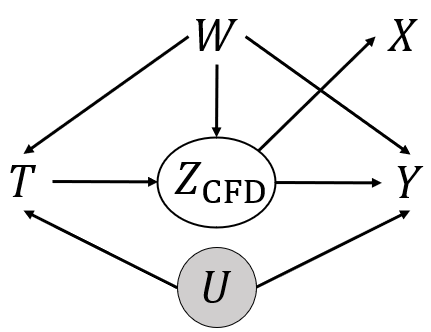}
	\caption{DAG $\mathcal{G}$ that represents the data generation mechanism assumed in this paper.}
	\label{pic:Problem}
	\vspace{-0.5cm}
\end{wrapfigure}

We assume data is generated based on the DAG $\mathcal{G}$ in Fig.~\ref{pic:Problem}, where $T$ is the treatment variable, $Y$ is the outcome variable, $U$ is the unobserved confounding variable, $X$ is the proxy of $Z_{\text{\rm CFD}}$, the latent CFD adjustment variable whose representation is to be learned and used for CFD adjustment, and $W$ is the observed confounding variable and is the conditioning variable of $Z_{\text{\rm CFD}}$.

In our problem setting, we assume that the observed confounding variable $W$ and the proxy variable $X$ are naturally separable. We believe this assumption is easy to satisfy in practice since $W$ is a pre-treatment variable (measured before treatment assignment) while $X$ is the proxy of the post-treatment variable, which is always collected after treatment assignment. For instance, with the example in the Introduction, as previously mentioned, $W$ can be a patient's age, and $X$ can be the results of some follow-up tests after the treatment has been applied, such as sputum and urine tests.

To clarify, the latent variable (i.e., $Z_{\text{\rm CFD}}$) refers to the variable that is not measured, but its information is captured by its proxy. On the other hand, the unobserved confounding variable (i.e., $U$) is not measured and has no proxy. Latent variables and the existence of their proxies are commonly assumed by data-driven causal inference methods~\citep{LouizosSMSZW17,ZhangLL21,cheng2022causal} and it is a practical assumption. In addition to the previous example where follow-up medical test results can be a proxy for tar in lungs, another example would be in the case when we are not able to measure a person's economic status, so a common solution is to rely on the proxy variable such as postcode~\citep{angrist2009mostly,montgomery2000measuring}.

We summarise the assumptions and the goal of the CFDiVAE model as follows.
\begin{problem}
	Given a joint probability distribution $P(X,W,T,Y)$ that is generated from the underlying DAG in Fig.~\ref{pic:Problem} where $U$ and $Z_{\text{\rm CFD}}$ are not measured. Suppose that $X$ is the proxy of the latent variable $Z_{\text{\rm CFD}}$. The goal of CFDiVAE is to learn the  representation of $Z_{\text{\rm CFD}}$.
\end{problem}

For the simplicity of notation and without causing confusion, in the rest of the paper, we use $Z_{\text{\rm CFD}}$ to represent the learned representation of the latent variable $Z_{\text{\rm CFD}}$ in Fig.~\ref{pic:Problem}, unless otherwise stated.

\subsection{Representation Learning}
In this section, we introduce the details of CFDiVAE for learning $Z_{\text{\rm CFD}}$. CFDiVAE learns a full generative model $p(X, Z_{\text{\rm CFD}} \mid T,W) = p(X \mid Z_{\text{\rm CFD}})p(Z_{\text{\rm CFD}} \mid T,W)$ and an inference model $q(Z_{\text{\rm CFD}} \mid T,W,X)$.

To guarantee the identifiability of CFDiVAE, we take $T$ and $W$ as additionally observed variables to approximate the prior $p(Z_{\text{\rm CFD}} \mid T, W)$~\citep{KhemakhemKMH20}. Following existing VAE-based works in~\citep{LouizosSMSZW17,ZhangLL21,cheng2022causal}, we assume the prior $p(Z_{\text{\rm CFD}} \mid T, W)$ follows the Gaussian distribution, that is:
\begin{equation}
	\begin{aligned}
		p(Z_{\text{\rm CFD}} \mid T, W) = \prod_{j=1}^{D_{{Z_{\text{\rm CFD}}}}}\mathcal{N}({Z_{\text{\rm CFD}}}_j \mid \mu = 0, \sigma^2 =1),
	\end{aligned}
\end{equation}where $D_{{Z_{\text{\rm CFD}}}}$ is the dimension of ${Z_{\text{\rm CFD}}}$.

In the inference model, we design the encoder $q(Z_{\text{\rm CFD}} \mid T,W,X)$ that serves as the variational approximation of the posterior over the target representation, and the variational approximation of the posterior is defined as follows:
\begin{equation}
	\begin{aligned}
		q(Z_{\text{\rm CFD}} \mid T,W,X) = \prod_{j=1}^{D_{Z_{\text{\rm CFD}}}}\mathcal{N}(\mu = \hat{\mu}_{{Z_{\text{\rm CFD}}}_j}, \sigma^2 = \hat{\sigma}^2_{{Z_{\text{\rm CFD}}}_j}),
	\end{aligned}
\end{equation}where $\hat{\mu}_{{Z_{\text{\rm CFD}}}}$ and $\hat{\sigma}^2_{{Z_{\text{\rm CFD}}}}$ are the means and variances of the Gaussian distributions parameterised by the neural networks for $Z_{\text{\rm CFD}}$. 

The generative model for $X$ is defined as: 
\begin{equation}
	\begin{aligned}
		p(X \mid {Z_{\text{\rm CFD}}}) = \prod_{j=1}^{D_{X}}\mathcal{N}(X_j\mid\mu = \hat{\mu}_{X_{j}}, \sigma^2 = \hat{\sigma}^2_{X_{j}});~ \hat{\mu}_{X_{j}} = g({Z_{\text{\rm CFD}}});~\hat{\sigma}^2_{X_{j}} = g({Z_{\text{\rm CFD}}}),
	\end{aligned}
\end{equation}where $D_{{X}}$ is the dimension of ${X}$, and $g(\cdot)$ is a neural network parameterised by its own parameters. 

Then the evidence lower bound (ELBO) for the above inference and generative models is as follows:
\begin{equation}
	\begin{aligned}
		\mathcal{M}_\text{CFDiVAE} =~ &\mathbb{E}_{q}[\log p(X \mid Z_{\text{\rm CFD}})] - D_{\mathrm{KL}}[q(Z_{\text{\rm CFD}} \mid T,W,X)~||~p(Z_{\text{\rm CFD}} \mid T, W)],
		\label{eqa:ELBO}
	\end{aligned}
\end{equation}where $D_{\mathrm{KL}}[\cdot||\cdot]$ is a $\mathrm{KL}$ divergence term. 

\subsection{Model Identifiability Analysis}
In this section, we provide the identifiability analysis of our model. CFDiVAE is identifiable if the following implication holds.
\begin{equation}
	\begin{aligned}
		\forall (\bm{\theta},\bm{\theta'}): p_{\bm{\theta}}(X, Z_{\text{\rm CFD}} \mid T,W) = p_{\bm{\theta'}}(X, Z_{\text{\rm CFD}} \mid T,W) \Longrightarrow \bm{\theta} = \bm{\theta'}
	\end{aligned}
\end{equation}

Let $\bm{\theta} = (\ervf,\ermS, \bm{\lambda})$ be the parameters of the following conditional generative model:
\begin{equation}
	\label{eq:001}
	\begin{aligned}
		p_{\bm{\theta}}(X, Z_{\text{\rm CFD}} \mid T,W) = p_{\ervf}(X \mid Z_{\text{\rm CFD}})p_{\ermS, \bm{\lambda}}(Z_{\text{\rm CFD}} \mid T,W),
	\end{aligned}
\end{equation}and we define: 
\begin{equation}
	\label{eq:002}
	\begin{aligned}
		p_{\ervf}(X \mid Z_{\text{\rm CFD}}) = p_{\bm{\varepsilon}}(X - \ervf(Z_{\text{\rm CFD}})).
	\end{aligned}
\end{equation}
This means that the value of $X$ can be decomposed as $X = \ervf(Z_{\text{\rm CFD}}) + \bm{\varepsilon}$, where $\bm{\varepsilon}$ is an independent noise variable with probability density function $p_{\bm{\varepsilon}}(\bm{\varepsilon})$. However, our model also applies to non-noisy proxy variable and in this case $X = \ervf(Z_{\text{\rm CFD}})$. We assume that the function $\ervf$ is injective. 

For the prior $p_{\ermS, \bm{\lambda}}(Z_{\text{\rm CFD}} \mid T,W)$, we have the following assumption, i.e., conditionally factorial, where each element of $Z_{\text{\rm CFD}}$ has an exponential family distribution given $T$ and $W$. 
\begin{assumption}
	We assume that the probability density function is given by:
	\begin{equation}
		\label{eq:003}
		\begin{aligned}
			p_{\ermS, \bm{\lambda}}(Z_{\text{\rm CFD}} \mid T,W) = \prod_{i}^{D_{{Z_{\text{\rm CFD}}}}}\frac{Q_i({Z_{\text{\rm CFD}}}_i)}{{Z}_i(T, W)}exp\left[ \sum_{j=1}^{k}S_{i,j}({Z_{\text{\rm CFD}}}_i)\lambda_{i,j}(T, W)\right],
		\end{aligned}
	\end{equation}where $Q_i$ is the base measure, ${Z}_i(T, W)$ is the normalising constant and $\ermS_i = (S_{i,1},...,S_{i,k})$ are sufficient statistics and $\bm{\lambda}(T,W) =({\lambda}_{i,1}(T,W),...,{\lambda}_{i,k}(T,W)) $ are the corresponding parameters depending on $T$ and $W$, and $k$ is the dimension of each sufficient statistic.
\end{assumption}

Following the work in~\citep{KhemakhemKMH20}, let $X \in \mathbb{R}^{d}$ and $Z_{\text{\rm CFD}} \in \mathbb{R}^{n}$ ($n \leq d$), we have the following theorem about the identifiability of our model.
\begin{theorem}
	\label{def:identiable}
	Assume that the observational data are generated according to Eq.~\ref{eq:001}-Eq.~\ref{eq:003} with parameters $\bm{\theta} = (\ervf,\ermS, \bm{\lambda})$ and the following hold: (1) The function $\ervf$ in Eq.~\ref{eq:002} is injective. (2) The set $\{ X \in \mathcal{X} \mid \varphi_{\ervf} (X) =0\}$ has measure zero, where $\varphi_{\bm{\varepsilon}}$ is the characteristic function of the density $p_{\bm{\varepsilon}}$ defined in Eq.~\ref{eq:002}. (3) The sufficient statistics $S_{i,j}$ in Eq.~\ref{eq:003} are differentiable almost everywhere, and $(S_{i,j})_{1\leq j \leq k}$ are linearly independent on any subset of $\mathcal{X}$ of measure greater than zero. (4) There exists $nk+1$ distinct points $(T,W)_{0},...,(T,W)_{nk}$ such that the matrix $\mL = (\bm{\lambda}(T_1,W_1)-\bm{\lambda}(T_0,W_0),...,\bm{\lambda}(T_{nk},W_{nk})-\bm{\lambda}(T_0,W_0))$ of size $nk \times nk$ is invertible. Then the parameters $\bm{\theta} = (\ervf,\ermS, \bm{\lambda})$ are $\thicksim_{\mA}$-identifiable.
\end{theorem}

This theorem guarantees the identifiability of the generative model in Eq.~\ref{eq:001}. Proof of the theorem is provided in Appx.~\ref{sec:proofi} and more related definitions are available in Appx.~\ref{Identifiability}.

\section{ATE Estimation}
After learning ${Z_{\text{\rm CFD}}}$, we can obtain unbiased estimation of the ATE by using the CFD adjustment. In the following, we show how this is done with data generated under a linear model. For data generated under a nonlinear model, we refer readers to the literature, e.g., \citep{tchetgen2012semiparametric} since this step (estimating ATE using a given adjustment variable) is beyond our contribution.

For the following linear model, 
\begin{equation*}
	\begin{aligned}
		 {Z_{\text{\rm CFD}}} &= c_{{Z_{\text{\rm CFD}}}} + \beta_{T,{Z_{\text{\rm CFD}}}}T + \beta_{W,{Z_{\text{\rm CFD}}}}W + e_{{Z_{\text{\rm CFD}}}};\\
		 Y &= c_{Y} + \beta_{Y,{{Z_{\text{\rm CFD}}}}}{Z_{\text{\rm CFD}}} + \beta_{W,Y}W + \beta_{U,Y}U + e_{Y},
	\end{aligned}
\end{equation*}where $c$ denotes intercept and $e$ denotes error, the ATE of $T$ on $Y$ is the product of coefficients $\beta_{T,Z_{\text{\rm CFD}}}$ and $\beta_{Y,Z_{\text{\rm CFD}}}$. The coefficients are obtained with the following process~\citep{barr_2018}: 
\begin{enumerate}
	\item ${Z_{\text{\rm CFD}}}$ is regressed on $T$ and $W$. This gives us the coefficient $\beta_{T,{Z_{\text{\rm CFD}}}}$ and $\mathbb{E}[{Z_{\text{\rm CFD}}}\mid T,W]$. Using $\mathbb{E}[{Z_{\text{\rm CFD}}}\mid T,W]$, we estimate the noise $e_{Z_{\text{\rm CFD}}}$ as ${Z_{\text{\rm CFD}}} - \mathbb{E}[{Z_{\text{\rm CFD}}}\mid T,W]$.
	
	\item Regress $e_{Z_{\text{\rm CFD}}}$ on $Y$. This gives us the coefficient $\beta_{Y,Z_{\text{\rm CFD}}}$. Noise $e_{Z_{\text{\rm CFD}}}$ is only introduced at ${Z_{\text{\rm CFD}}}$, and is independent of the unobserved confounding variable $U$.
\end{enumerate}

\section{Experiments}
In this section, we first demonstrate the correctness of representation learning. Then, we compare the performance of CFDiVAE with the benchmark methods for estimating causal effects and validate that CFDiVAE can unbiasedly estimate the causal effects and its performance is not sensitive to the change of the causal strength of the unobserved confounding variable. We also show its feasibility when the dimension of the learned representation is mismatched with the dimension of the ground truth CFD adjustment variable. Finally, we apply CFDiVAE to a real-world dataset and demonstrate its potential application. We also provide an additional experiment on the analysis of model identifiability in Appx.~\ref{sec:Analysis}. The source code is available in the Supplementary Material.

\subsection{Experiment Setup}
\begin{wraptable}{r}{0.5\textwidth}
	\vspace{-0.5cm}
	\caption{Methods for comparison.}
	\label{tab:models}
	\centering
	\setlength\tabcolsep{1pt}
	{\small  \begin{tabular}{cc}
			\toprule
			Name 		& Open-Source          \\ \midrule
			LinearDRL	\citep{chernozhukov2018double}	& \href{https://econml.azurewebsites.net/_autosummary/econml.dml.LinearDML.html}{EconML} \\
			CausalForest	\citep{wager2018estimation}	& \href{https://econml.azurewebsites.net/_autosummary/econml.dml.CausalForestDML.html}{EconML} \\
			ForestDRL	\citep{athey2019generalized}	& \href{https://econml.azurewebsites.net/_autosummary/econml.dr.ForestDRLearner.html}{EconML} \\
			XLearner	\citep{kunzel2019metalearners}	& \href{https://econml.azurewebsites.net/_autosummary/econml.metalearners.XLearner.html}{EconML} \\
			KernelDML	\citep{nie2021quasi}	& \href{https://econml.azurewebsites.net/_autosummary/econml.dml.KernelDML.html}{EconML} \\
			CEVAE	\citep{LouizosSMSZW17}	& \href{https://github.com/AMLab-Amsterdam/CEVAE}{GitHub} \\
			TEDVAE	\citep{ZhangLL21}	& \href{https://github.com/WeijiaZhang24/TEDVAE}{GitHub} \\ \bottomrule
	\end{tabular}}
	\vspace{-0.3cm}
\end{wraptable}

We compare CFDiVAE with a number of benchmark methods, including traditional and VAE based causal effect estimation methods, as listed in Table~\ref{tab:models}. The implementations of CEVAE and TEDVAE are retrieved from the authors’ GitHub and the implementations of other methods are from EconML~\citep{econml}. The detailed description of the comparison methods is shown in Appx.~\ref{sec:Description}. 

F{\scriptsize IND}FDS{\scriptsize ET} and L{\scriptsize IST}FDS{\scriptsize ETS}~\citep{jeong2022finding,wienobst2022finding} are the only existing front-door adjustment based methods. They are not selected for comparison since they require a known DAG, which is often not available. Moreover, it is not possible to learn the underlying DAG from data in our case due to the unobserved confounding variable.

For evaluating the performance of CFDiVAE and the benchmark methods, we use the Estimation Bias $|(\hat{\beta} - \beta)/\beta|\times100\%$ as the metric, where $\hat{\beta}$ is the estimated ATE and $\beta$ is the ground truth.  

The evaluation of estimated causal effects with unobserved confounding variables relies on synthetic datasets since no ground truth causal effects available for real-world datasets~\citep{LouizosSMSZW17,ZhangLL21,cheng2022causal}. Synthetic datasets used in the evaluation are generated based on the causal graph (mechanism) shown in Fig~\ref{pic:Problem}. More details on data generation are provided in the Supplementary Material. To avoid the bias brought by the data generation process, we repeatedly generate 30 datasets with a range of sample sizes (denoted as $\text{N}$), including 0.5k, 1k, 2k, 4k, 6k, 8k, 10k and 20k. For each method, we report the average (mean) estimation bias over the 30 datasets, together with the standard deviation.

\subsection{Correctness of the learned representation} 
\begin{wrapfigure}{r}{0.45\textwidth}
	\vspace{-0.6cm}
	\centering
	\includegraphics[scale=0.3]{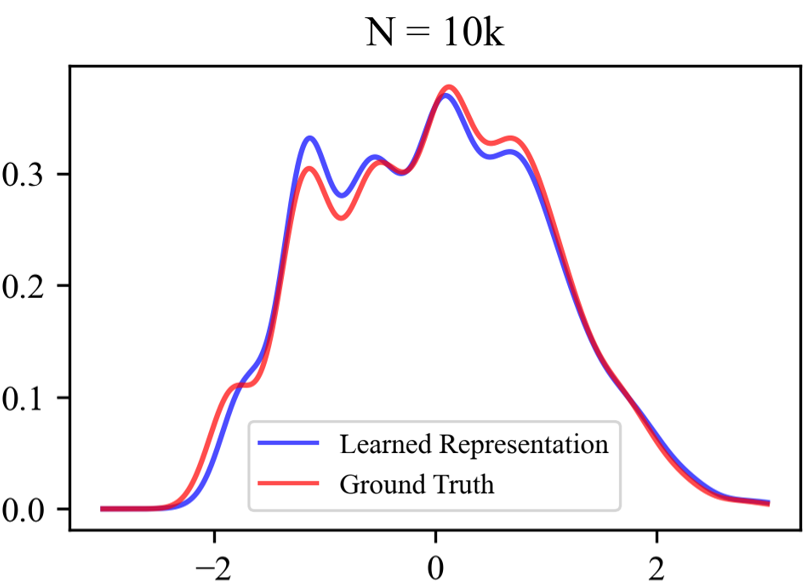}
	\caption{Probability Density Functions of the ground truth and the learned representation, where the horizontal axis represents the value and the vertical axis represents the density.}
	\label{pic:correct}
	\vspace{-0.8cm}
\end{wrapfigure}

In this section, we conduct experiments to validate the correctness of the learned representation. Since we use synthetic datasets, we know the ground truth of the CFD adjustment variable. To evaluate the correctness of the representations learned by CFDiVAE, we compare the probability distribution of the learned representation against the distribution of the corresponding ground truth CFD adjustment variable. Due to page limit, we only show the result of $\text{N}$=10k. As shown in Fig.~\ref{pic:correct}, the distribution of the learned representation is close to the distribution of the ground truth, which indicates that CFDiVAE can learn accurate representation of the CFD adjustment variable. More results are reported in Appx.~\ref{sec:correctapp}.

\subsection{Performance of ATE Estimation}
In this section, we evaluate the performance of CFDiVAE in ATE estimation compared with the benchmark methods. As shown in Table~\ref{tab:syn}, CFDiVAE outperforms all the other comparison methods when the sample size is 2k and above. Such results are expected. All comparison methods use the back-door adjustment to estimate ATE, i.e., they use $W$ as the back-door adjustment variable. The estimation bias for comparison methods is due to the unobserved confounding variable $U$. To obtain unbiased estimation based on the back-door adjustment, all back-door paths between $T$ and $Y$ must be blocked, but this is impossible as the back-door path via $U$ cannot be blocked because $U$ is unobserved. Our proposed method CFDiVAE circumvents the limitations of back-door adjustment. 

\begin{table}[t]
	\vspace{-0.8cm}
	\setlength\tabcolsep{3pt}
	\caption{The estimation bias (\%) of CFDiVAE and comparison methods under different $N$ values.}
	\label{tab:syn}
	\centering
	{\scriptsize   \begin{tabular}{ccccccccc}
			\toprule
			& 0.5k          & 1k            & 2k            & 4k           & 6k           & 8k           & 10k          & 20k          \\ \midrule
			LinearDRL    & 21.90 ± 5.13   & 21.56 ± 3.82  & 21.47 ± 3.28  & 21.82 ± 2.08 & 21.59 ± 1.78 & 21.88 ± 1.41 & 21.89 ± 1.31 & 21.38 ± 0.90  \\
			CausalForest & 21.87 ± 5.55  & 21.33 ± 4.28  & 21.39 ± 3.62  & 21.85 ± 1.98 & 21.63 ± 1.80  & 21.88 ± 1.33 & 21.94 ± 1.23 & 21.36 ± 0.99 \\
			ForestDRL    & 21.90 ± 4.95   & 21.58 ± 3.69  & 21.36 ± 3.38  & 21.79 ± 2.04 & 21.54 ± 1.80  & 21.88 ± 1.38 & 21.89 ± 1.28 & 21.41 ± 0.89 \\
			XLearn       & 21.92 ± 5.14  & 21.65 ± 3.55  & 21.35 ± 3.36  & 21.83 ± 2.04 & 21.59 ± 1.78 & 21.86 ± 1.39 & 21.88 ± 1.30  & 21.39 ± 0.90  \\
			KernelDML    & 19.57 ± 5.38  & 19.63 ± 3.83  & 19.79 ± 3.56  & 20.38 ± 2.04 & 20.24 ± 1.75 & 20.59 ± 1.39 & 20.64 ± 1.25 & 20.27 ± 0.94 \\
			CEVAE        & 102.63 ± 2.83 & 104.31 ± 7.82 & 101.42 ± 20.50 & 31.05 ± 4.95 & 26.93 ± 5.04 & 23.97 ± 6.05 & 21.29 ± 6.81 & 28.83 ± 4.72 \\
			TEDVAE       & 98.91 ± 17.37 & 70.73 ± 16.94 & 26.67 ± 3.58  & 24.63 ± 2.28 & 22.84 ± 1.85 & 22.67 ± 1.61 & 22.63 ± 1.23 & 21.84 ± 0.98 \\ \midrule
			CFDiVAE        & 86.29 ± 6.21  & 39.72 ± 31.47 & 8.87 ± 10.68  & 4.57 ± 3.03  & 2.58 ± 1.96  & 2.32 ± 1.47  & 2.97 ± 2.09  & 2.14 ± 3.38  \\ \bottomrule
	\end{tabular}}
\vspace{-0.5cm}
\end{table}

\subsection{Impact of the Causal Strength of Unobserved Confounding Variable}
We also conduct experiments to verify the effectiveness of CFDiVAE with respect to different causal strengths of the unobserved confounding variable. For this set of experiments, the causal strength is varied by adjusting the coefficient of the path $U \to Y$. The sample size for this experiment is fixed at 10k. We multiply a scaling factor to the coefficient (i.e., $\beta_{U,Y}$) to realise the different causal strength levels of the unobserved confounding variable. For example, $0.0$ means that there is no unobserved confounding variable, and $2.0$ means that the coefficient doubles the original value. The range of the scaling factor is set as $[0.0, 2.0]$ and the step increment is set as $0.2$. 

\begin{figure}[t]
	\vspace{-0.8cm}
	\centering
	\includegraphics[scale=0.45]{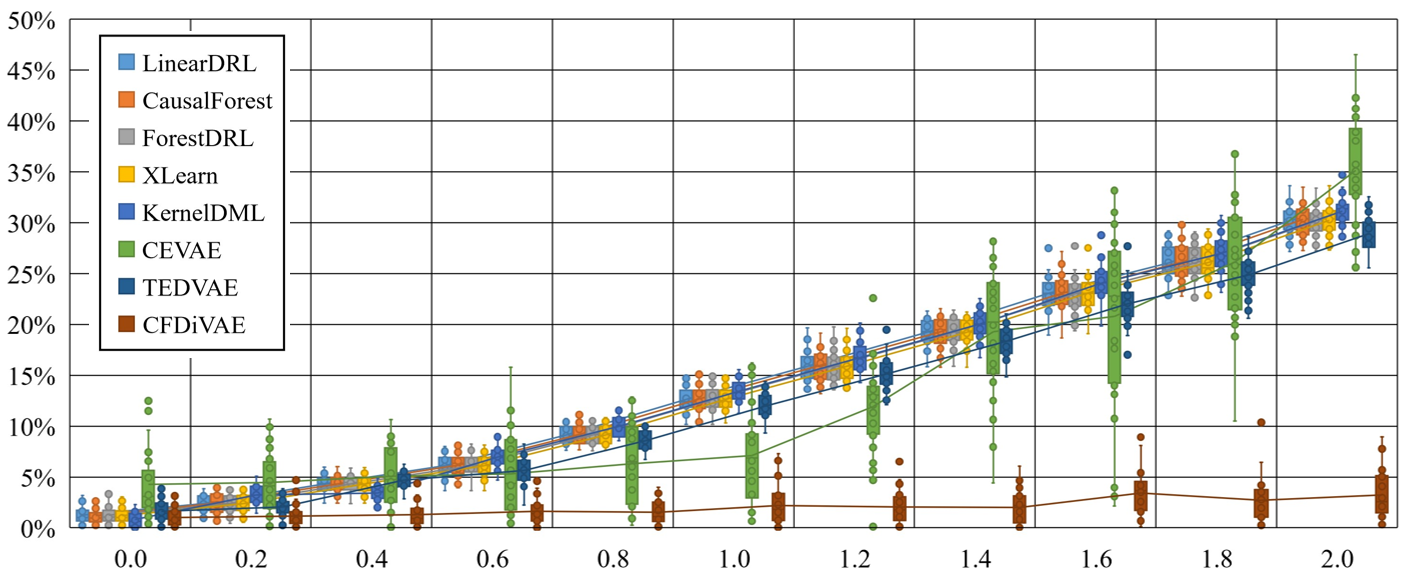}
	\caption{Results with different scaling factor, where the horizontal axis represents the scaling factor and the vertical axis represents the estimation bias (\%).}
	\label{pic:RES_U}
\end{figure}

The results are shown in Fig.~\ref{pic:RES_U}. When the causal strength is zero, i.e., no unobserved confounding variable, the comparison methods each achieve their own best performance since in this case, all confounding variables are observed and their performance is solely determined by their capabilities in correctly identifying or learning the correct back-door adjustment variable. With the increase of causal strength, there is a clear downward trend in the performance of the comparison methods, indicating that the back-door adjustment cannot handle unobserved confounding variable. In contrast, CFDiVAE achieves and maintains an estimation bias of around $3\%$. The result is expected as CFDiVAE is based on the CFD adjustment, which is able to cope with unobserved confounding variables.

\subsection{Sensitivity to Representation Dimension} 
In real-world applications, it is a common situation that the dimension set for the representation does not match the dimension of the ground truth CFD adjustment variable. In this section, we analyse the sensitivity of CFDiVAE to representation dimension. In the following, $D_{\text{L}}$ represents the dimension of the learned representation, while $D_{\text{R}}$ represents the dimension of the ground truth CFD adjustment variable. We apply CFDiVAE to various dimension settings, i.e., $D_{\text{R}} \in \{2,4,8\}$. The results are shown in Table~\ref{tab:DIM}. We see that CFDiVAE achieves its best performance when $D_{\text{L}} = D_{\text{R}}$. When $D_{\text{L}} \neq D_{\text{R}}$, the performance of CFDiVAE can also maintain at an acceptable level. In all cases, the performance of CFDiVAE is superior to the comparison methods (Appx.~\ref{sec:dimensionality} shows more results). Hence, when the dimension of the ground truth CFD adjustment variable is not accessible, we can safely set $D_{\text{L}} = 1$.    

\begin{table}[t]
	\setlength\tabcolsep{3pt}
	\caption{The estimation bias ($\%$) of CFDiVAE to dimension mismatch on different $\text{N}$ values. The best results are shown in \textbf{boldface}.}
	\label{tab:DIM}
	\centering
	{\scriptsize\begin{tabular}{ccccccccc}
			\toprule
			$D_{\text{L}}$-$D_{\text{R}}$ & 0.5k & 1k & 2k & 4k & 6k     & 8k & 10k & 20k \\ \midrule
			1-2 & 82.31 ± 8.83           & \textbf{11.99 ± 5.98}  & 10.7 ± 17.07           & 9.52 ± 3.08           & 9.54 ± 2.34          & 9.86 ± 2.54          & 10.35 ± 4.25         & 9.88 ± 1.36          \\
			2-2 & \textbf{78.16 ± 4.99}  & 12.85 ± 10.96          & \textbf{6.90 ± 5.88}    & \textbf{8.83 ± 6.02}  & \textbf{5.94 ± 4.22} & \textbf{5.46 ± 3.62} & \textbf{5.37 ± 6.82} & \textbf{4.16 ± 8.90}  \\ \midrule
			1-4 & 79.94 ± 8.98           & 22.12 ± 18.63          & 12.09 ± 4.62           & 13.73 ± 3.58          & 14.24 ± 3.43         & 15.07 ± 2.86         & 14.33 ± 2.64         & 14.83 ± 1.74         \\
			2-4 & 74.31 ± 6.90            & \textbf{16.38 ± 8.4}   & \textbf{9.49 ± 5.02}   & 11.54 ± 3.75          & 9.84 ± 3.15          & 8.19 ± 4.86          & 8.43 ± 6.85          & 6.10 ± 1.83           \\
			4-4 & \textbf{73.16 ± 5.70}   & 19.04 ± 11.12          & 12.89 ± 16.47          & \textbf{9.9 ± 5.15}   & \textbf{8.74 ± 5.69} & \textbf{6.78 ± 3.92} & \textbf{4.50 ± 2.70}   & \textbf{4.45 ± 1.75} \\ \midrule
			1-8 & 75.38 ± 12.60           & 27.92 ± 22.61          & 15.86 ± 6.77           & 14.46 ± 17.05         & 15.18 ± 4.85         & 16.62 ± 3.73         & 16.64 ± 3.34         & 17.65 ± 2.29         \\
			4-8 & 72.33 ± 11.25          & \textbf{19.45 ± 13.21} & \textbf{12.89 ± 10.46} & 11.71 ± 12.01         & 12.28 ± 5.88         & 10.36 ± 5.37         & 9.26 ± 5.88          & 7.47 ± 3.04          \\
			8-8 & \textbf{63.95 ± 11.41} & 27.47 ± 13.96          & 29.00 ± 26.66             & \textbf{11.01 ± 11.10} & \textbf{10.00 ± 7.61}   & \textbf{9.00 ± 6.27}    & \textbf{6.69 ± 4.29} & \textbf{7.42 ± 3.14} \\ \bottomrule
	\end{tabular}}
	\vspace{-0.3cm}
\end{table}

\subsection{Case Study on A Real-World Dataset}
In this section, we apply CFDiVAE to detect discrimination on the real-world dataset, Adult. The dataset is retrieved from the UCI repository~\citep{asuncion2007uci} and it contains 11 attributes about personal, educational and economic information for 48842 individuals. We use the sensitive attribute sex as $T$, income as $Y$, and age, race and $\text{native}\_\text{country}$ as $W$. We consider all the other attributes such as $\text{maritial}\_\text{status}$ as the proxy of the latent CFD adjustment variable, which represents the stereotype held by the society, and it is the stereotype that directly causes discrimination.

With causality-based discrimination detection, we consider that there is direct discrimination if the sensitive attribute has a large enough direct causal effect on the outcome (above a given threshold $\tau$), and there is indirect discrimination if the sensitive attribute has a large enough indirect causal effect on the outcome and the mediator is also a sensitive attribute~\citep{zhang2018causal}.

With the Adult dataset, ~\cite{ZhangWW17} found that there was no direct discrimination but significant indirect discrimination against sex through the indirect paths via
$\text{marital}\_\text{status}$ ($\tau = 0.05$). When we apply CFDiVAE to the Adult dataset and use the learned representation for CFD adjustment, the estimated average causal effect of sex on income is $0.176$, indicating significant discrimination against sex through the stereotype. The evaluation is consistent with the conclusion shown in~\citep{ZhangWW17}. More details and explanations of this case study are reported in Appx.~\ref{sec:case}.

\section{Related Work}
Over the past few decades, researchers have proposed many methods for estimating causal effects from observational data. These methods generally fall into three categories: methods based on back-door adjustment, instrumental variables (IVs) and front-door adjustment, respectively.

Methods based on back-door adjustment~\citep{pearl2009causality} are the most widely used, and most of these methods need to assume that all confounding variables are observed. For example, several tree-based models~\citep{athey2016recursive,su2009subgroup,zhang2017mining} have been designed to estimate causal effects by designing specific splitting criteria; meta-learning~\citep{kunzel2019metalearners} has also been proposed to utilise existing machine learning algorithms for causal effect estimation. Recently, methods using deep learning techniques to predict causal effects have received widespread attention. For example, CEVAE~\citep{LouizosSMSZW17} combines representation learning and VAE to estimate causal effects; TEDVAE~\citep{ZhangLL21} improves on CEVAE and decouples the learned representations to achieve more accurate estimation; Counterfactual regression nets~\citep{johansson2016learning,shalit2017estimating,hassanpour2019counterfactual} balances treated and untreated sample groups so that the two groups are as close as possible.

Methods based on IVs have also received a lot of attention. Most IV based methods require users to nominate a valid IV, such as the generalised method of moments (GMM)~\citep{bennett2019deep}, kernel-IV regression~\citep{singh2019kernel} and deep learning based method~\citep{hartford2017deep}. When there are no nominated IVs in the data, some data-driven methods are developed to find~\citep{yuan2022auto} or synthesise~\citep{burgess2013use,kuang2020ivy} an IV or eliminate the influence of invalid IVs by using statistical strategies~\citep{guo2018confidence,hartford2021valid}. 

Front-door adjustment based approach is rarely studied in the literature. There are only a few methods for finding appropriate adjustment sets by following the standard front-door criterion~\citep{jeong2022finding,wienobst2022finding}. These methods require a given DAG and aim to find and enumerate possible standard front-door adjustment variables in the DAG. 

The methods based on back-door adjustment cannot handle unobserved confounding variables. IV based methods can cope with unobserved confounding variables, but the availability of known IVs is itself a strong assumption. Existing front-door adjustment based methods all require a given DAG and a standard front-door adjustment variable, both of which are often difficult to obtain in practice. We propose the CFD adjustment to relax the restriction of standard front-door adjustment and develop CFDiVAE to learn a CFD adjustment variable for unbiased ATE estimation in the presence of unobserved confounding variables.

\section{Conclusion}
\textbf{Summary of Contributions.} This work studies the practical and challenging problem of causal effect estimation from observational data when there exist unobserved confounding variables. We have proposed the conditional front-door adjustment, which is less restrictive than the standard front-door adjustment and proved that average causal effect is identifiable via the proposed conditional front-door adjustment. Our proposed  CFDiVAE model leverages the identifiable VAE technique to learn the representation of the conditional front-door adjustment variable from data directly without assuming a given causal graph, and we have shown that the identifiability of the learned representation is theoretically guaranteed. Extensive experiments have demonstrated that CFDiVAE outperforms the benchmark methods. We have also shown that CFDiVAE is insensitive to the causal strength of the unobserved confounding variable. Furthermore, the case study has suggested the potential of CFDiVAE for real-world applications. In summary, our work provides a novel and more practical approach to causal effect estimation from observation data with unobserved confounders.

\textbf{Limitations \& Future Works.} The success of the proposed conditional front-door adjustment and CFDiVAE model relies on some assumptions. Although these assumptions are common in causal inference research or VAE-based models, there may still be situations where the assumptions cannot be satisfied. In future, we will investigate how to relax these assumptions to further improve the opportunities for using causal inference to solve real-world problems.

\bibliography{CFDiVAE}
\bibliographystyle{iclr2024_conference}

\newpage
\appendix

\section{Background}
\subsection{Causality}
\label{sec:Causality}
In a DAG (directed acyclic graph) $\mathcal{G}=(\mathbf{V}, \mathbf{E})$, a path $\pi$ between nodes $V_{1}$ and $V_{n}$ comprises a sequence of distinct nodes $<V_{1}, \dots, V_{n}>$ with every pair of successive nodes being adjacent. A node $V$ lies on the path $\pi$ if $V$ belongs to the sequence $<V_{1}, \dots, V_{n}>$. 

A path $\pi$ is causal if all edges along it are all in the same direction such as $V_{1} \to ... \to V_{n}$. A path that is not causal is referred to as a non-causal path.

\begin{definition}[$d$-separation~\citep{pearl2009causality}]
	\label{d-separation}
	A path $\pi$ in a DAG is said to be $d$-separated (or blocked) by a set of nodes ${Z}$ iff (1) the path $\pi$ contains a chain $V_i \rightarrow V_k \rightarrow V_j$ or a fork $V_i \leftarrow V_k \rightarrow V_j$ such that the middle node $V_k$ is in ${Z}$, or (2) the path $\pi$ contains an inverted fork (or collider) $V_i \rightarrow V_k \leftarrow V_j$ such that $V_k$ is not in ${Z}$ and no descendant of $V_k$ is in ${Z}$.
\end{definition}

Let $\mathcal{G}=(\mathbf{V}, \mathbf{E})$ be a DAG, and $P(V)$ is the probability distribution over $V$. In the DAG $\mathcal{G}$, a set of nodes ${Z}$ is said to $d$-separate $V_i$ and $V_j$ if and only if ${Z}$ blocks every path between $V_i$ to $V_j$; otherwise, a set of nodes ${Z}$ is said to $d$-connect $V_i$ and $V_j$. When the Markov Condition and Faithfulness assumption are satisfied by $\mathcal{G}$ and $P(V)$, $(V_i \indep V_j \mid {Z})$ if ${Z}$ $d$-separates $V_i$ and $V_j$, and $(V_i \nindep V_j \mid {Z})$ if ${Z}$ $d$-connects $V_i$ and $V_j$.

\begin{theorem}[Rules of $do$-Calculus~\citep{pearl2009causality}]
	 \label{do}
	 Let $\mathcal{G}$ be the DAG associated with a causal model, and let $P(\cdot)$ stand for the probability distribution induced by that model. For any disjoint subsets of variables $T,Y,Z$, and $W$, we have the following rules.
	
	~~~~Rule 1. (Insertion/deletion of observations):
	\begin{equation*}
		\begin{aligned}
			P(y \mid do(t),z,w) = P(y \mid do(t),w), ~\text{if}~(Y \indep Z \mid T,W)~\text{in}~\mathcal{G}_{\overline{T}}.
		\end{aligned}
	\end{equation*}
	~~~~Rule 2. (Action/observation exchange):
	\begin{equation*}
		\begin{aligned}
			P(y \mid do(t), do(z), w) = P(y \mid do(t),z, w), ~\text{if}~(Y\indep Z \mid T,W)~\text{in}~\mathcal{G}_{\overline{T}\underline{Z}}.
		\end{aligned}
	\end{equation*}
	~~~~Rule 3. (Insertion/deletion of actions):
	\begin{equation*}
		\begin{aligned}
			P(y \mid do(t),do(z), w) = P(y \mid do(t), w), ~\text{if}~(Y\indep Z \mid T, W)~\text{in}~\mathcal{G}_{\overline{T},\overline{Z(W)}}~\text{,}
		\end{aligned}
	\end{equation*}~~~~~~~~where $Z(W)$ is the nodes in $Z$ that are not ancestors of any node in $W$ in $\mathcal{G}_{\overline{T}}$.
\end{theorem}

\subsection{Model Identifiability}
\label{Identifiability}
We define two equivalence relations on the set of parameters $\Theta$.
\begin{definition}
	Let $\thicksim$ be the equivalence relation on $\Theta$ defined as follows:
	\begin{equation}
		\begin{aligned}
			(\ervf,\ermS, \bm{\lambda}) \thicksim_{\mA} (\widetilde{\ervf},\widetilde{\ermS},\widetilde{\bm{\lambda}}) \Leftrightarrow\\
			\exists\mA,\ervc \mid \ermS(\inv{\ervf}(\vx)) = \mA\widetilde{\ermS}(\inv{\widetilde{\ervf}}(\vx))+\ervc, \forall\vx\in \mathcal{X},
		\end{aligned}
	\end{equation}where $(\widetilde{\ervf},\widetilde{\ermS},\widetilde{\bm{\lambda}})$ are the parameters obtained from some learning algorithm that perfectly approximates the marginal distribution of the observations, $\mA$ is an invertible $nk \times nk$ matrix, $\ervc$ is a vector, and $\mathcal{X}$ is the domain of $X$.
\end{definition}

\section{Proofs}
\begin{figure}[t]
	\centering
	\includegraphics[scale=0.35]{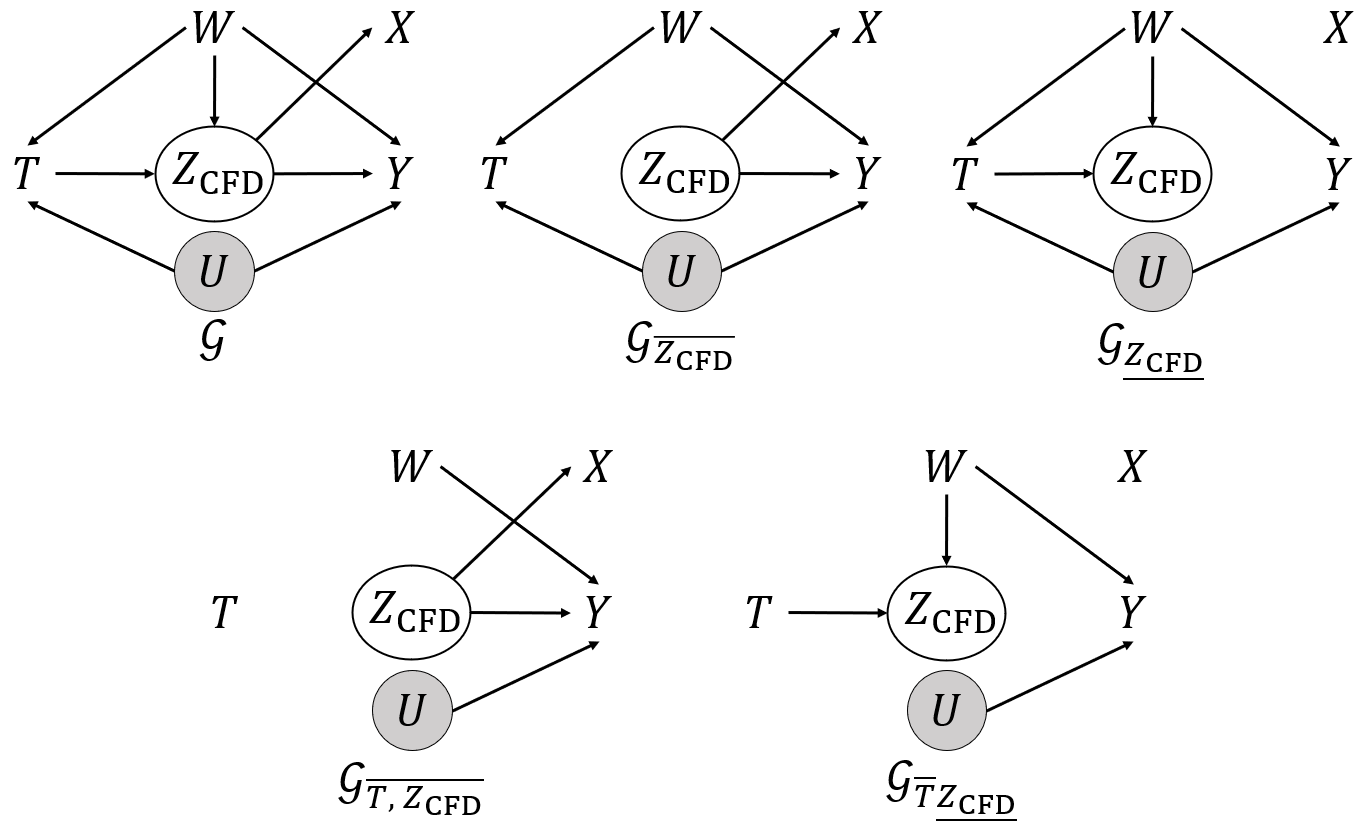}
	\caption{Subgraphs of $\mathcal{G}$ used in the derivation of causal effects.}
	\label{pic:proofs}
\end{figure}

\subsection{Proof of theorem~\ref{def:FD2}}
\label{sec:proofFD2}
\begin{proof}
	We compute $P(y \mid do(t))$ by using Theorem~\ref{do} under the DAG $\mathcal{G}$ in Fig.~\ref{pic:Problem}. Fig.~\ref{pic:proofs} shows the subgraphs that are needed for the derivations in the following. 
	
	$P(y \mid do(t))$ can be expanded as:
	\begin{equation}
		\begin{aligned}
			P(y \mid do(t)) = \sum_{z_{\text{\rm CFD}}}P(z_{\text{\rm CFD}}\mid do(t)) P(y\mid z_{\text{\rm CFD}},do(t)) \label{eq:014}
		\end{aligned}
	\end{equation}
	We first compute $P(y\mid z_{\text{\rm CFD}},do(t))$, which can be expanded as follow:
	\begin{align}
		& P(y\mid z_{\text{\rm CFD}},do(t)) = \sum_{w} P(y \mid do(t),z_{\text{\rm CFD}},w) P(w\mid do(t),z_{\text{\rm CFD}}) \\
		\text{The first part:}~& P(y \mid do(t),z_{\text{\rm CFD}},w) = P(y \mid do(t),do(z_{\text{\rm CFD}}),w), \nonumber \\
		&~~~~~~~~\text{ since }  (Y \indep Z_{\text{\rm CFD}}\mid T, W) \text{ in } \mathcal{G}_{\overline{T}\underline{Z_{\text{\rm CFD}}}} \text{ (Rule 2 in Theorem~\ref{do})} \nonumber \\
		& P(y \mid do(t),do(z_{\text{\rm CFD}}),w) = P(y \mid do(Z_{\text{\rm CFD}}),w), \nonumber \\
		&~~~~~~~~\text{ since }  (Y \indep T \mid Z_{\text{\rm CFD}}, W) \text{ in } \mathcal{G}_{\overline{Z_{\text{\rm CFD}}}\overline{T(W)}} \text{ (Rule 3 in Theorem~\ref{do})} \nonumber \\
		& P(y \mid do(z_{\text{\rm CFD}}),w) = \sum_{t'} P(y \mid do(z_{\text{\rm CFD}}),t',w) P(t' \mid do(z_{\text{\rm CFD}}),w) \nonumber \\
		& P(y \mid do(z_{\text{\rm CFD}}),t',w) = P(y \mid z_{\text{\rm CFD}},t',w), \nonumber \\
		&~~~~~~~~\text{ since }  (Y \indep Z_{\text{\rm CFD}} \mid T, W) \text{ in } \mathcal{G}_{\underline{Z_{\text{\rm CFD}}}} \text{ (Rule 2 in Theorem~\ref{do})} \nonumber \\
		& P(t' \mid do(z_{\text{\rm CFD}}),w) = P(t' \mid w), \nonumber \\
		&~~~~~~~~\text{ since }  (T \indep Z_{\text{\rm CFD}} \mid W) \text{ in } \mathcal{G}_{\overline{Z_{\text{\rm CFD}}(W)}} \text{ (Rule 3 in Theorem~\ref{do})} \nonumber \\
		& P(y \mid do(t),z_{\text{\rm CFD}},w) = \sum_{t'} P(y \mid t',z_{\text{\rm CFD}},w) P(t' \mid w) \\
		\text{The second part:}~& P(w\mid do(t),z_{\text{\rm CFD}}) = P(w,z_{\text{\rm CFD}} \mid do(t))/P(z_{\text{\rm CFD}} \mid do(t)) \nonumber \\
		& P(w,z_{\text{\rm CFD}} \mid do(t)) = P(z_{\text{\rm CFD}} \mid t, w)P(w) \nonumber \\
		& P(w\mid do(t),\vzfd) = \frac{P(z_{\text{\rm CFD}} \mid t, w)P(w)}{P(z_{\text{\rm CFD}} \mid do(t))} \\
		\text{Thus,}~& P(y\mid z_{\text{\rm CFD}},do(t)) = \sum_{w,t'} P(y \mid t',z_{\text{\rm CFD}},w) P(t' \mid w)\frac{P(z_{\text{\rm CFD}} \mid t, w)P(w)}{P(z_{\text{\rm CFD}} \mid do(t))} \label{eq:019}
	\end{align}
	We take Eq.~\ref{eq:019} into Eq.~\ref{eq:014} and get,
	\begin{align}
		P(y \mid do(t)) &= \sum_{z_{\text{\rm CFD}}}P(z_{\text{\rm CFD}}\mid do(t)) \sum_{w,t'} P(y \mid t',z_{\text{\rm CFD}},w) P(t' \mid w)\frac{P(z_{\text{\rm CFD}} \mid t, w)P(w)}{P(z_{\text{\rm CFD}} \mid do(t))}\nonumber \\
		&= \sum_{z_{\text{\rm CFD}},w,t'}P(z_{\text{\rm CFD}}\mid do(t)) P(y \mid t',z_{\text{\rm CFD}},w) P(t' \mid w)\frac{P(z_{\text{\rm CFD}} \mid t, w)P(w)}{P(z_{\text{\rm CFD}} \mid do(t))}
	\end{align}
	Finally, we get,
	\begin{align}
		P(y \mid do(t)) &= \sum_{z_{\text{\rm CFD}},w,t'} P(y \mid t',z_{\text{\rm CFD}},w) P(t' \mid w) P(z_{\text{\rm CFD}} \mid t,w) P(w)
	\end{align}where $t'$ is a distinct realisation of treatment.
\end{proof}

\subsection{Proof of theorem~\ref{def:identiable}}
\label{sec:proofi}
Our proof is based on the proof of Theorem 1 in~\citep{KhemakhemKMH20}.
\begin{proof}
	Suppose we have two sets of parameters $(\ervf,\ermS, \bm{\lambda})$ and $(\widetilde{\ervf},\widetilde{\ermS},\widetilde{\bm{\lambda}})$ such that $p_{\bm{\theta}}(X, Z_{\text{\rm CFD}} \mid T,W) = p_{\widetilde{\bm{\theta}}}(X, Z_{\text{\rm CFD}} \mid T,W)$. Then:
	\begin{align}
		&~~~~~~~~\int_{\mathcal{Z}_\text{CFD}} p_{\ermS, \bm{\lambda}}(Z_{\text{\rm CFD}} \mid T,W)p_{\ervf}(X \mid Z_{\text{\rm CFD}})\text{d}_{Z_{\text{\rm CFD}}} = \int_{\mathcal{Z}_\text{CFD}} p_{\widetilde{\ermS}, \widetilde{\bm{\lambda}}}(Z_{\text{\rm CFD}} \mid T,W)p_{\widetilde{\ervf}}(X \mid Z_{\text{\rm CFD}})\text{d}_{Z_{\text{\rm CFD}}} \nonumber\\
		&\Longrightarrow \int_{\mathcal{Z}_\text{CFD}} p_{\ermS, \bm{\lambda}}(Z_{\text{\rm CFD}} \mid T,W)p_{\bm{\varepsilon}}(X - \ervf(Z_{\text{\rm CFD}}))\text{d}_{Z_{\text{\rm CFD}}} = \int_{\mathcal{Z}_\text{CFD}} p_{\widetilde{\ermS}, \widetilde{\bm{\lambda}}}(Z_{\text{\rm CFD}} \mid T,W)p_{\bm{\varepsilon}}(X - \widetilde{\ervf}(Z_{\text{\rm CFD}}))\text{d}_{Z_{\text{\rm CFD}}} \nonumber\\
		&\Longrightarrow \int_{\mathcal{X}} p_{\ermS, \bm{\lambda}}(\inv{\ervf}(\bar{X}) \mid T,W)\text{vol}{J}_{\inv{\ervf}}(\bar{X})p_{\bm{\varepsilon}}(X - \bar{X})\text{d}_{\bar{X}} \nonumber\\
		&~~~~~~~~~~~~= \int_{\mathcal{X}} p_{\widetilde{\ermS}, \widetilde{\bm{\lambda}}}(\inv{\widetilde{\ervf}}(\bar{X}) \mid T,W)\text{vol}{J}_{\inv{\widetilde{\ervf}}}(\bar{X})p_{\bm{\varepsilon}}(X - \bar{X})\text{d}_{\bar{X}} \label{eq:30}
	\end{align}
We denote the volume of a matrix $\text{vol}~\mA$, and when $\mA$ is full column rank, $\text{vol}~\mA = \sqrt{\text{det}\mA^{T}\mA}$. $J$ denotes the Jacobian, and we make the change of the variable $\bar{X} = \ervf(Z_{\text{\rm CFD}})$ on the left hand side, and $\bar{X} = \widetilde{\ervf}(Z_{\text{\rm CFD}})$ on the right hand side.

From Eq.~\ref{eq:30}, we have:
\begin{align}
	{p}_{\ermS, \bm{\lambda}}(\inv{\ervf}(\bar{X}) \mid T,W)\text{vol}{J}_{\inv{\ervf}}(\bar{X}) = {p}_{\widetilde{\ermS}, \widetilde{\bm{\lambda}}}(\inv{\widetilde{\ervf}}(\bar{X}) \mid T,W)\text{vol}{J}_{\inv{\widetilde{\ervf}}}(\bar{X}) \label{eq:40}
\end{align}
By taking the logarithm on the both sides of Eq.~\ref{eq:40} and replacing $p_{\ermS, \bm{\lambda}}$ by its expression from Eq.~\ref{eq:003}, we get:
\begin{align}
	\text{log}~\text{vol}{J}_{\inv{\ervf}}({X}) + \sum_{i=1}^{n}(\text{log}~Q_i(\inv{f_i}(X)) - \text{log}~{Z}_i(T, W) + \sum_{j=1}^{k}S_{i,j}(\inv{f_i}(X))\lambda_{i,j}(T, W)) = \nonumber \\
	\text{log}~\text{vol}{J}_{\inv{\widetilde{\ervf}}}({X}) + \sum_{i=1}^{n}(\text{log}~\widetilde{Q}_i(\inv{\widetilde{f}_i}(X)) - \text{log}~\widetilde{{Z}}_i(T, W) + \sum_{j=1}^{k}\widetilde{S}_{i,j}(\inv{\widetilde{f}_i}(X))\widetilde{\lambda}_{i,j}(T, W)) \label{eq:50}
\end{align}
Let $(T,W)_{0},...,(T,W)_{nk}$ be the points provided by Theorem~\ref{def:identiable}~(4), and define $\bar{\bm{\lambda}}(T,W) = \bm{\lambda}(T,W) - \bm{\lambda}(T_0,W_0)$. We plug each of those $(T,W)_{l}$ in Eq.~\ref{eq:50} to obtain $nk+1$ such equations. We subtract the first equation for $(T,W)_{0}$ from the remaining $nk$ equations to get for $l = 1,...,nk$:
\begin{align}
	\langle \ermS({\inv{\ervf}}({X})),\bar{\lambda}(T_l,W_l)\rangle + \sum_{i}\text{log}~\frac{{Z}_i(T_0, W_0)}{{Z}_i(T_l, W_l)} = \langle \widetilde{\ermS}({\inv{\widetilde{\ervf}}}({X})),\bar{\widetilde{\lambda}}(T_l,W_l)\rangle + \sum_{i}\text{log}~\frac{\widetilde{{Z}}_i(T_0, W_0)}{\widetilde{{Z}}_i(T_l, W_l)} \label{eq:60}
\end{align}
Let $\mL$ be the matrix defined in Theorem~\ref{def:identiable}~(4), and $\widetilde{\mL}$ similarly defined for $\widetilde{\bm{\lambda}}$ ($\widetilde{\mL}$ is not necessarily invertible). Define $b_l = \sum_{i}\text{log}~\frac{\widetilde{{Z}}_i(T_0, W_0){Z}_i(T_l, W_l)}{{Z}_i(T_0, W_0)\widetilde{{Z}}_i(T_l, W_l)}$ and $\ervb$ the vector of all $b_l$ for $l = 1,...,nk$.

Then, Eq.~\ref{eq:60} can be rewritten as:
\begin{align}
	\mL^{T}\ermS({\inv{\ervf}}({X})) = \widetilde{\mL}^{T}\widetilde{\ermS}({\inv{\widetilde{\ervf}}}({X})) + \ervb \label{eq:70}
\end{align}
We multiply both sides of Eq.~\ref{eq:70} by the transpose of the inverse of $\mL^{T}$ from the left to get:
\begin{align}
	\ermS({\inv{\ervf}}({X})) = \mA\widetilde{\ermS}({\inv{\widetilde{\ervf}}}({X})) + \ervc , \label{eq:80}
\end{align}where $\mA = \mL^{-T}\widetilde{\mL}$ and $\ervc = \mL^{-T}\ervb$.

By definition of $\ermS$ and according to Theorem~\ref{def:identiable}~(3), its Jacobian exists and is an $nk\times n$ matrix of rank $n$. This implies that the Jacobian of $\widetilde{\ermS} \circ {\inv{\ervf}}$ exists and is of rank $n$ and so is $\mA$. We have two cases: (1) If $k = 1$, $\mA$ is invertible since $\mA$ is $n \times n$ matrix of rank $n$; (2) If $k >= 2$, $\mA$ is also invertible. We have the following proof for (2): 

Define $\bar{X} = {\inv{\ervf}}({X})$ and $\ermS_i(\bar{X}_i) = (S_{i,1}(\bar{X}_i),...,S_{i,k}(\bar{X}_i))$. For each $i \in [1,...,n]$ there exist $k$ points $\bar{X}_i^{1},...,\bar{X}_i^{k}$ such that $(\ermS_{i}^{'}(\bar{X}_i^1),...,\ermS_{i}^{'}(\bar{X}_i^k))$ are linearly independent. 

Firstly, we proof the above statement. Suppose that for any choice of such $k$ points, the family $(\ermS_{i}^{'}(\bar{X}_i^1),...,\ermS_{i}^{'}(\bar{X}_i^k))$ is never linearly independent. That means that $\ermS_{i}^{'}(\mathbb{R})$ is included in a subspace of $\mathbb{R}^{k}$ of dimension at most $k-1$. Let $\vh$ a non zero vector that is orthogonal to $\ermS_{i}^{'}(\mathbb{R})$. Then for all $X \in \mathbb{R}$, we have $\langle \ermS_{i}^{'}(\mathbb{R}),\vh \rangle = 0$. By integrating we find that $\langle \ermS_{i}(\mathbb{R}),\vh \rangle = \text{const}$. Since this is true for all $X \in \mathbb{R}$ and for a $\vh \neq 0$, we conclude that the distribution is not strongly exponential, which contradicts our hypothesis. 

Secondly, we prove $\mA$ is invertible. Collect those points into $k$ vectors $(\bar{X}^{1},...,\bar{X}^{k})$, and concatenate the $k$ Jacobians $J_{\ermS}(\bar{X}^{l})$ evaluated at each of those vectors
horizontally into the matrix $\mQ = (J_{\ermS}(\bar{X}^{1}),...,J_{\ermS}(\bar{X}^{k}))$ (and similarly define $\widetilde{\mQ}$ as the concatenation of the Jacobians of $\widetilde{\ermS}({\inv{\widetilde{\ervf}}}\circ{\ervf}(\bar{X}))$ evaluated at those points). Then the matrix $\mQ$ is invertible. By differentiating Eq.~\ref{eq:80} for each $X^{l}$, we have:
\begin{align}
	\mQ = \mA \widetilde{\mQ}
\end{align}
The invertibility of $\mQ$ implies the invertibility of $\mA$ and $\widetilde{\mQ}$, which completes the proof.
\end{proof}

\section{Experiment}
\subsection{Description of the Comparison Methods}
\label{sec:Description}
\textbf{LinearDRL}~\cite{chernozhukov2018double}: A double machine learning estimator with a low-dimensional linear regression as the final stage.

\textbf{CausalForest}~\cite{wager2018estimation}: A causal forest estimator combined with the double machine learning technique for conditional average treatment effect estimation.

\textbf{ForestDRL}~\cite{athey2019generalized}: A generalised random forest and orthogonal random forest  based estimator that uses doubly-robust correction techniques to account for covariates shift (or selection bias) between the treatment.

\textbf{XLearner}~\cite{kunzel2019metalearners}: A meta-learning algorithm that utilises supervised learning methods (e.g., Random Forests and Bayesian Regression) for the analysis of conditional average treatment effects.

\textbf{KernelDML}~\cite{nie2021quasi}: A specialised version of the double machine learning estimator that uses random fourier features and kernel ridge regression for the analysis of conditional average treatment effects.

\textbf{CEVAE}~\cite{LouizosSMSZW17}: A deep learning based method that leverages latent variable modelling, specifically Variational AutoEncoder, to estimate causal effect from observational data, even in the presence of latent confounders.

\textbf{TEDVAE}~\cite{ZhangLL21}: A deep learning based method that learns the disentangled representations of confounding, instrumental, and risk factors using VAE for accurate treatment effect estimation.

\subsection{More Results of the Experiments in Section 4.2}
\label{sec:correctapp}
\begin{figure}[h]
	\centering
	\includegraphics[scale=0.6]{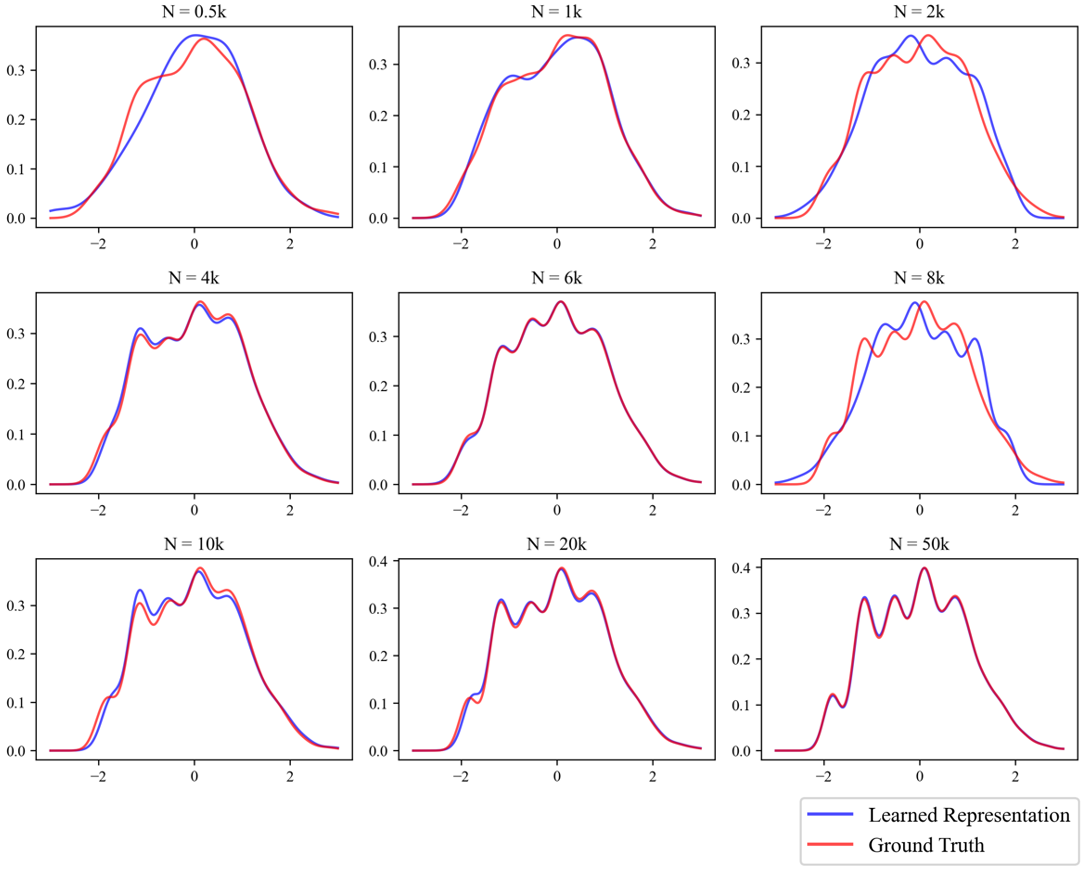}
	\caption{Probability Density Functions of the ground truth CFD adjustment variable and the learned representation, where the horizontal axis represents the value and the vertical axis represents the density.}
	\label{pic:correctapp}
\end{figure}

In this section, we compare the probability distribution of the learned representation of the CFD adjustment variable with the distribution of the ground truth CFD adjustment variable under different sample sizes. As shown in Fig.~\ref{pic:correctapp}, the distribution of the learned representation is close to the ground truth distribution, which indicates that the proposed method CFDiVAE can learn the accurate representation of the CFD adjustment variable from its proxy.

\subsection{Analysis of Model Identifiability}
\label{sec:Analysis}
Our proposed model CFDiVAE takes $T$ and $W$ as additional observed variables to approximate the prior $p(Z_{\text{\rm CFD}} \mid T, W)$. In this section, we apply two partially identifiable VAE models, i.e., $T$-CFDiVAE and $W$-CFDiVAE, and the original VAE as comparison methods. $T$-CFDiVAE is partially identifiable VAE model that takes $T$ as the additional observed variable to approximate $p(Z_{\text{\rm CFD}} \mid T)$; $W$-CFDiVAE is partially identifiable VAE model that takes $W$ as the additional observed variable to approximate $p(Z_{\text{\rm CFD}} \mid W)$; the original VAE does not take any additional observed variable to approximate $p(Z_{\text{\rm CFD}})$. The ELBOs for these models are defined as: 
\begin{align}
	\mathcal{M}_{T\text{-CFDiVAE}} &=~\mathbb{E}_{q}[\log p(X \mid Z_{\text{\rm CFD}})] - D_{\mathrm{KL}}[q(Z_{\text{\rm CFD}} \mid T,X)~||~p(Z_{\text{\rm CFD}} \mid T)] \label{eq:tFDiVAE}\\
	\mathcal{M}_{W\text{-CFDiVAE}} &=~\mathbb{E}_{q}[\log p(X \mid Z_{\text{\rm CFD}})] - D_{\mathrm{KL}}[q(Z_{\text{\rm CFD}} \mid W,X)~||~p(Z_{\text{\rm CFD}} \mid W)] \label{eq:wFDiVAE} \\
	\mathcal{M}_{\text{VAE}} &=~\mathbb{E}_{q}[\log p(X \mid Z_{\text{\rm CFD}})] - D_{\mathrm{KL}}[q(Z_{\text{\rm CFD}} \mid X)~||~p(Z_{\text{\rm CFD}})] \label{eq:VAE}
\end{align}
The results are shown in Table~\ref{tab:ablation}. We see that CFDiVAE achieves the best performance since it uses all additional observed variables. The performance of $T$-CFDiVAE is slightly lower than the performance of $W$-CFDiVAE since $W$ has more additional information than $T$ (the dimension of $W$ is generally higher than the dimension of $T$). The original VAE which does not use any additional observed variable achieves the worst performance. 

\begin{table}[t]
	\setlength\tabcolsep{2pt}
	\caption{Results of model identifiability analysis.}
	\label{tab:ablation}
	\centering
	{\scriptsize \begin{tabular}{ccccccccc}
			\toprule
			& 0.5k          & 1k          & 2k         & 4k          & 6k          & 8k         & 10k       & 20k       \\ \midrule
			$\mathcal{M}_{\text{VAE}}$~(Eq.~\ref{eq:VAE})     & 88.93 ± 14.67  & 58.41 ± 22.46 & 77.71 ± 9.17 & 79.89 ± 7.00     & 32.92 ± 31.20  & 10.49 ± 7.66 & 7.07 ± 5.72 & 4.55 ± 9.19 \\
			$\mathcal{M}_{T\text{-CFDiVAE}}$~(Eq.~\ref{eq:tFDiVAE}) & 70.09 ±   20.60 & 77.48 ± 12.65 & 81.48 ± 8.19 & 15.94 ± 22.84 & 10.25 ± 22.99 & 4.60 ± 6.11   & 3.89 ± 4.56 & 1.77 ± 1.41 \\
			$\mathcal{M}_{W\text{-CFDiVAE}}$~(Eq.~\ref{eq:wFDiVAE}) & 68.89 ± 18.89  & 85.83 ± 4.64  & 29.13 ± 21.00   & 5.49 ± 3.83   & 6.49 ± 17.82  & 3.14 ± 2.04  & 3.14 ± 2.35 & 1.69 ± 1.55 \\ \midrule
			$\mathcal{M}_\text{CFDiVAE}$~(Eq.~\ref{eqa:ELBO})  & 86.29 ± 6.21   & 39.72 ± 31.47 & 8.87 ± 10.68 & 4.57 ± 3.03   & 2.58 ± 1.96   & 2.32 ± 1.47  & 2.97 ± 2.09 & 1.57 ± 1.32 \\ \bottomrule
	\end{tabular}}
	\vspace{-0.3cm}
\end{table}

\subsection{More Results of the Experiments in Section 4.5}
\label{sec:dimensionality}
In this section, we evaluate the performance of CFDiVAE and the comparison methods when the dimension for the representation does not match the dimension of the ground truth CFD adjustment variable. Table~\ref{tab:Res2} shows the results for $D_{\text{R}} = 2$, Table~\ref{tab:Res4} shows the results for $D_{\text{R}} = 4$, and Table~\ref{tab:Res8} shows the results for $D_{\text{R}} = 8$. We note that CFDiVAE achieves its best performance when $D_{\text{L}} = D_{\text{R}}$ and the performance of CFDiVAE is better than the comparison methods even when the dimensiona for the representation is fixed at 1. Hence, in a more general case, when the dimension of the ground truth CFD adjustment variable is not accessible, we can safely set $D_{\text{L}} = 1$ to get an acceptable causal effect estimation.

\begin{table}[h]
	\setlength\tabcolsep{2pt}
	\caption{The estimation bias (\%) of CFDiVAE and comparison methods under different $N$ values. CFDiVAE-$D_{\text{L}}$-$D_{\text{R}}$ denotes apply CFDiVAE to a specified setting, where $D_{\text{L}}$ represents the dimension of the learned representation and $D_{\text{R}}$ represents the dimension of the ground truth CFD adjustment variable.}
	\label{tab:Res}
	\centering
	\begin{subtable}[h]{0.99\textwidth}
		\centering
		\caption{Estimation bias ($\%$) when $D_{\text{R}} = 2$.}
		\label{tab:Res2}
		{\scriptsize \begin{tabular}{ccccccccc}
			\toprule
			& 0.5k          & 1k            & 2k             & 4k           & 6k           & 8k            & 10k           & 20k          \\ \midrule
			LinearDRL    & 27.05 ± 7.56  & 24.58 ± 6.39  & 26.13 ± 4.19   & 24.53 ± 3.01 & 25.01 ± 2.90  & 24.92 ± 1.61  & 25.40 ± 1.56   & 25.42 ± 1.12 \\
			CausalForest & 28.26 ± 8.21  & 24.51 ± 6.76  & 26.01 ± 4.48   & 24.56 ± 3.11 & 25.09 ± 3.10  & 24.98 ± 1.58  & 25.40 ± 1.67   & 25.46 ± 1.13 \\
			ForestDRL    & 26.91 ± 7.95  & 24.51 ± 6.38  & 26.20 ± 3.99    & 24.54 ± 3.11 & 24.95 ± 2.87 & 24.96 ± 1.60   & 25.38 ± 1.59  & 25.43 ± 1.13 \\
			XLearn       & 27.15 ± 7.50   & 24.62 ± 6.29  & 26.04 ± 3.94   & 24.57 ± 3.09 & 25.02 ± 2.90  & 24.94 ± 1.59  & 25.37 ± 1.56  & 25.40 ± 1.11  \\
			KernelDML    & 24.34 ± 7.84  & 22.01 ± 6.29  & 24.23 ± 4.09   & 22.94 ± 3.05 & 23.57 ± 2.81 & 23.57 ± 1.47  & 24.06 ± 1.58  & 24.22 ± 1.08 \\
			CEVAE        & 102.05 ± 3.22 & 104.47 ± 9.79 & 104.04 ± 22.15 & 41.27 ± 7.16 & 39.88 ± 8.89 & 32.34 ± 10.46 & 23.62 ± 11.21 & 34.20 ± 6.46  \\
			TEDVAE       & 93.05 ± 14.19 & 69.77 ± 21.04 & 24.61 ± 11.13  & 29.14 ± 2.81 & 26.55 ± 2.99 & 25.98 ± 1.50   & 26.22 ± 1.55  & 26.03 ± 1.30  \\ \midrule
			CFDiVAE-1-2    & 82.31 ± 8.83  & 11.99 ± 5.98  & 10.70 ± 17.07   & 9.52 ± 3.08  & 9.54 ± 2.34  & 9.86 ± 2.54   & 10.35 ± 4.25  & 9.88 ± 1.36  \\
			CFDiVAE-2-2    & 78.16 ± 4.99  & 12.85 ± 10.96 & 6.90 ± 5.88     & 8.83 ± 6.02  & 5.94 ± 4.22  & 5.46 ± 3.62   & 5.37 ± 6.82   & 4.16 ± 8.90   \\ \bottomrule
		\end{tabular}}
	\vspace{0.5cm}
	\end{subtable}
	\begin{subtable}[h]{0.99\textwidth}
		\centering
		\caption{Estimation bias ($\%$) when $D_{\text{R}} = 4$.}
		\label{tab:Res4}
		{\scriptsize \begin{tabular}{ccccccccc}
				\toprule
				& 0.5k          & 1k             & 2k            & 4k            & 6k            & 8k            & 10k           & 20k           \\ \toprule
				LinearDRL    & 32.82 ± 13.77 & 30.76 ± 10.62  & 32.91 ± 7.69  & 32.38 ± 4.51  & 32.38 ± 3.69  & 33.04 ± 3.33  & 32.13 ± 3.30   & 31.97 ± 1.98  \\
				CausalForest & 32.40 ± 13.57  & 31.34 ± 11.30   & 33.41 ± 7.79  & 32.88 ± 4.25  & 32.49 ± 3.94  & 33.18 ± 3.16  & 32.11 ± 3.36  & 31.94 ± 1.93  \\
				ForestDRL    & 32.42 ± 13.57 & 31.21 ± 10.75  & 32.78 ± 8.07  & 32.28 ± 4.31  & 32.43 ± 3.76  & 32.92 ± 3.20   & 32.13 ± 3.47  & 32.01 ± 1.97  \\
				XLearn       & 32.88 ± 12.93 & 31.14 ± 10.54  & 32.81 ± 7.65  & 32.36 ± 4.37  & 32.36 ± 3.69  & 32.98 ± 3.33  & 32.15 ± 3.35  & 32.00 ± 1.99     \\
				KernelDML    & 28.12 ± 14.57 & 27.83 ± 10.53  & 29.48 ± 7.97  & 30.47 ± 4.24  & 30.32 ± 3.72  & 31.19 ± 3.33  & 30.48 ± 3.17  & 30.60 ± 1.98   \\
				CEVAE        & 102.10 ± 2.45  & 102.03 ± 13.14 & 123.90 ± 39.63 & 34.56 ± 28.24 & 65.52 ± 16.65 & 50.36 ± 15.35 & 35.41 ± 17.62 & 42.41 ± 10.94 \\
				TEDVAE       & 99.99 ± 16.35 & 75.63 ± 40.63  & 30.72 ± 21.25 & 44.30 ± 4.87   & 35.52 ± 3.88  & 35.33 ± 3.24  & 33.82 ± 3.44  & 33.12 ± 2.20   \\ \midrule
				CFDiVAE-1-4    & 79.94 ± 8.98  & 22.12 ± 18.63  & 12.09 ± 4.62  & 13.73 ± 3.58  & 14.24 ± 3.43  & 15.07 ± 2.86  & 14.33 ± 2.64  & 14.83 ± 1.74  \\
				CFDiVAE-2-4    & 74.31 ± 6.90   & 16.38 ± 8.40    & 9.49 ± 5.02   & 11.54 ± 3.75  & 9.84 ± 3.15   & 8.19 ± 4.86   & 8.43 ± 6.85   & 6.10 ± 1.83    \\
				CFDiVAE-4-4    & 73.16 ± 5.70   & 19.04 ± 11.12  & 12.89 ± 16.47 & 9.90 ± 5.15    & 8.74 ± 5.69   & 6.78 ± 3.92   & 4.50 ± 2.70     & 4.45 ± 1.75   \\ \bottomrule
		\end{tabular}}
	\vspace{0.5cm}
	\end{subtable}
	\begin{subtable}[h]{0.99\textwidth}
		\centering
		\caption{Estimation bias ($\%$) when $D_{\text{R}} = 8$.}
		\label{tab:Res8}
		{\scriptsize \begin{tabular}{ccccccccc}
				\toprule
				& 0.5k          & 1k             & 2k             & 4k             & 6k            & 8k             & 10k           & 20k           \\ \midrule
				LinearDRL    & 54.02 ± 30.87 & 48.22 ± 16.82  & 47.84 ± 13.83  & 48.01 ± 9.29   & 46.89 ± 7.52  & 47.17 ± 4.97   & 48.66 ± 6.26  & 47.56 ± 3.31  \\
				CausalForest & 51.85 ± 31.93 & 49.64 ± 17.88  & 47.48 ± 13.26  & 47.06 ± 9.55   & 47.65 ± 7.71  & 47.68 ± 5.13   & 48.85 ± 6.41  & 47.47 ± 3.54  \\
				ForestDRL    & 53.42 ± 31.29 & 47.92 ± 17.51  & 47.76 ± 13.55  & 48.46 ± 9.34   & 46.74 ± 7.53  & 47.20 ± 4.83    & 48.60 ± 6.19   & 47.59 ± 3.34  \\
				XLearn       & 53.26 ± 30.78 & 48.08 ± 17.18  & 47.74 ± 13.99  & 48.06 ± 9.23   & 46.88 ± 7.56  & 47.15 ± 4.88   & 48.61 ± 6.23  & 47.54 ± 3.28  \\
				KernelDML    & 46.03 ± 31.34 & 43.16 ± 17.91  & 42.93 ± 13.83  & 44.79 ± 9.85   & 43.50 ± 7.86   & 44.59 ± 4.81   & 46.30 ± 6.26   & 45.65 ± 3.21  \\
				CEVAE        & 101.71 ± 2.29 & 107.12 ± 13.68 & 122.72 ± 46.09 & 106.34 ± 69.06 & 94.41 ± 29.52 & 106.92 ± 26.61 & 92.49 ± 36.07 & 33.48 ± 17.93 \\
				TEDVAE       & 98.58 ± 22.10  & 74.62 ± 55.07  & 60.77 ± 37.72  & 80.72 ± 11.28  & 59.95 ± 7.74  & 51.93 ± 5.16   & 52.78 ± 6.30   & 49.85 ± 3.43  \\ \midrule
				CFDiVAE-1-8    & 75.38 ± 12.60  & 27.92 ± 22.61  & 15.86 ± 6.77   & 14.46 ± 17.05  & 15.18 ± 4.85  & 16.62 ± 3.73   & 16.64 ± 3.34  & 17.65 ± 2.29  \\
				CFDiVAE-4-8    & 72.33 ± 11.25 & 19.45 ± 13.21  & 12.89 ± 10.46  & 11.71 ± 12.01  & 12.28 ± 5.88  & 10.36 ± 5.37   & 9.26 ± 5.88   & 7.47 ± 3.04   \\
				CFDiVAE-8-8    & 63.95 ± 11.41 & 27.47 ± 13.96  & 29.00 ± 26.66     & 11.01 ± 11.10   & 10.00 ± 7.61     & 9.00 ± 6.27       & 6.69 ± 4.29   & 7.42 ± 3.14   \\ \bottomrule
		\end{tabular}}
	\end{subtable}
\end{table}

\subsection{Explanations of the Case Study in Section 4.6}
\label{sec:case}
\begin{figure}[h]
	\centering
	\includegraphics[scale=0.42]{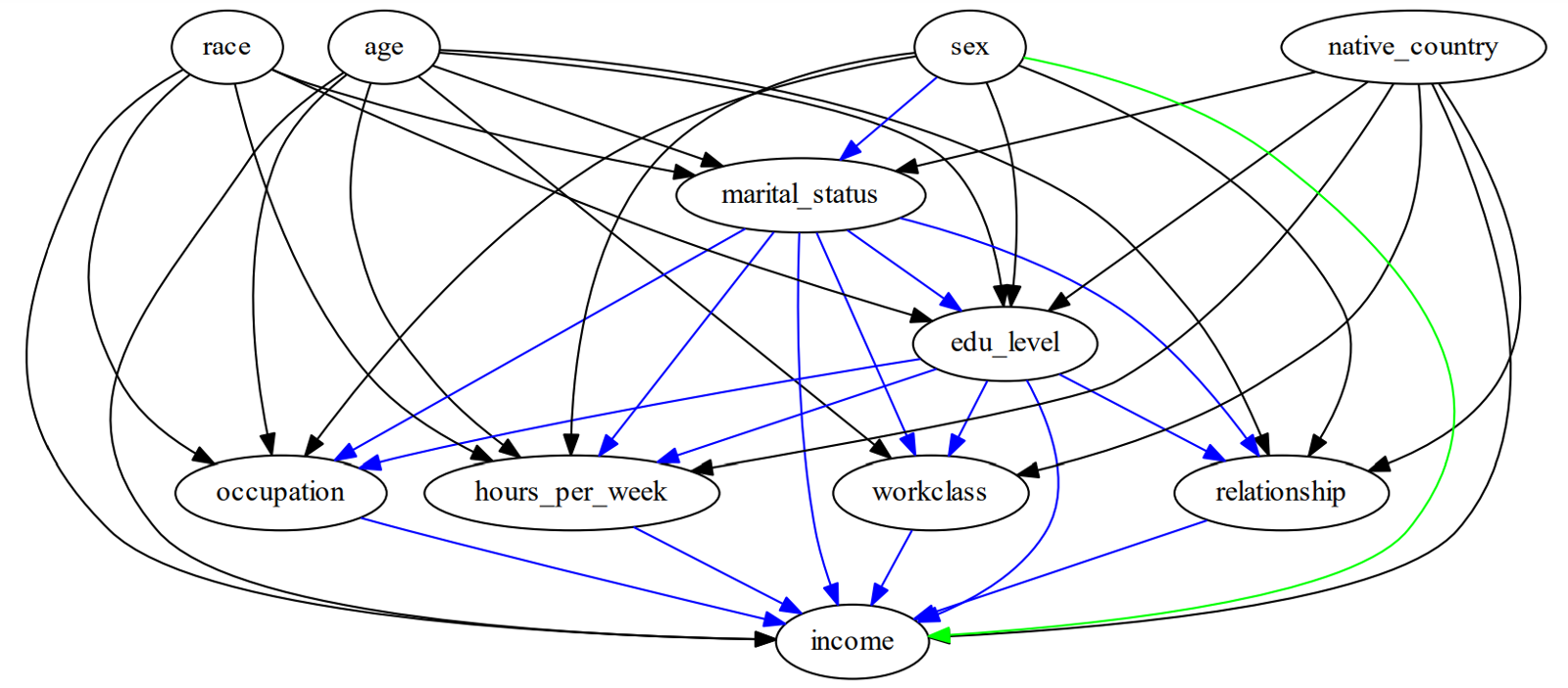}
	\caption{The causal network for the Adult dataset: the green path represents the direct path, and the blue paths represent the indirect paths passing through $\text{marital}\_\text{status}$~\citep{ZhangWW17}.}
	\label{pic:case}
\end{figure}

\begin{figure}[h]
	\centering
	\includegraphics[scale=0.36]{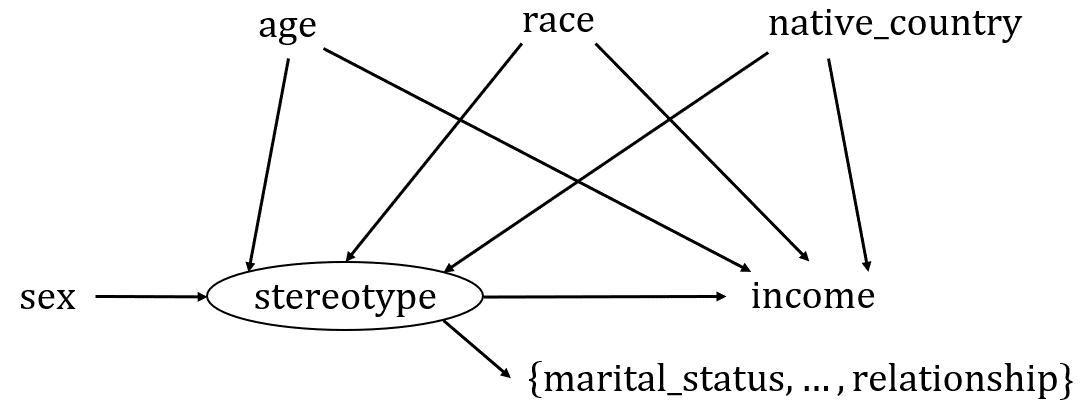}
	\caption{Simplified DAG for Adult dataset.}
	\label{pic:case2}
\end{figure}

Following the causal network in work~\citep{ZhangWW17}, the green
path represents the direct path from sex to income, and
the blue paths represent the indirect paths passing through
$\text{marital}\_ \text{status}$. The discrimination threshold $\tau$ is set as 0.05. By computing the path-specific effects, \cite{ZhangWW17} obtain that direct effect = 0.025 and indirect effect = 0.175, which indicate no direct discrimination but significant indirect discrimination.  

We aim to estimate the causal effect of sex on income using representation learning and conditional front-door adjustment. We simplify the above causal network to fit our model with a latent stereotype as shown in Fig.~\ref{pic:case2}. The discrimination is not a direct result of sex, but a direct result of the stereotype. The proxy of the stereotype is accessible, and they are $\text{marital}\_ \text{status}$, relationships, $\text{edu}\_ \text{level}$, $\text{hours}\_ \text{per}\_ \text{week}$, occupation, workclass, in this example. Stereotype is not a standard front-door adjustment variable because the causal path from stereotype to income is not blocked by sex. However, the stereotype is a CFD adjustment variable, since there are no back-door paths from sex to stereotype and all back-door paths from stereotype to income are blocked by age, race and $\text{native}\_ \text{country}$ (adding sex into this adjustment set will not invalidate the result). 

By using CFDiVAE, we obtain that $\text{ATE} = 0.176$, which is consistent with the previous estimate (0.175). The direct effect is ignored since it is very small. 

\section{Reproducibility}
In this section, we provide more details of the experimental setting and configuration for reproducibility purposes. CFDiVAE is implemented in Python~\citep{van1995python} libraries PyTorch~\citep{NEURIPS2019_9015} and Pyro~\citep{bingham2019pyro}. The code for data generation is written in R~\citep{R}. We provide the parameter settings of CFDiVAE in Table~\ref{tab:setting}. The descriptions of the major parameters are provided below:

\begin{itemize}
	\item Reps: the number of replications each set of experiments runs.
	\item Epoch: one Epoch is when an entire dataset is passed forward and backward through the neural network once.
	\item Batch\_Size: the number of training examples present in a
	single batch.
	\item Num\_Layers: the number of hidden layers.
	\item lr: the learning rate.
	\item lrd: the learning decay.
	\item wd: the weight decay.
\end{itemize}

\begin{table}[h]
	\centering
	\caption{Details of the parameter settings in FDVAE.}
	\label{tab:setting}
	\begin{tabular}{|cc|cc|cc|}
		\toprule
		Parameter & Value & Parameter & Value & Parameter & Value \\ \midrule
		Reps & 30 & Num\_Layers & 3 & wd & 1e-4 \\
		Epoch & 30 & lr & 1e-3 &  &  \\
		Batch\_Size & 256 & lrd & 0.01 &  &  \\ \bottomrule
	\end{tabular}
\end{table}

\end{document}